\documentclass[conference]{IEEEtran}
\usepackage{times}

\usepackage[numbers,sort&compress]{natbib}
\usepackage{multicol}
\usepackage{multirow}
\usepackage[bookmarks=true]{hyperref}
\usepackage{amsmath}
\usepackage{amsthm}
\usepackage{amssymb}
\usepackage{graphicx}
\usepackage[dvipsnames,table,xcdraw]{xcolor}
\usepackage[inkscapelatex=false]{svg}
\usepackage{caption}
\usepackage{subcaption}
\usepackage{graphicx}
\usepackage{soul}
\usepackage{cancel}
\usepackage[font=small,labelfont=bf]{caption}
\usepackage{physics}
\usepackage[normalem]{ulem}

\usepackage{soul}

\makeatletter

\newcommand{\revise}[1]{#1}

\makeatother
\newcommand\hcancel[2][black]{\setbox0=\hbox{$#2$}%
\rlap{\raisebox{.45\ht0}{\textcolor{#1}{\rule{\wd0}{1pt}}}}#2}

\newcommand{\nadiainline}[1]{#1}

\definecolor{mypink1}{rgb}{0.858, 0.188, 0.478}
\definecolor{mypink2}{RGB}{219, 48, 122}
\definecolor{mypink3}{cmyk}{0, 0.7808, 0.4429, 0.1412}
\definecolor{mygray}{gray}{0.6}

\usepackage{color}

\usepackage{lipsum} 
\usepackage{xcolor}
\hypersetup{
    colorlinks,
    linkcolor={red!50!black},
    citecolor={blue!50!black},
    urlcolor={blue!80!black}
}
\newcommand\xleftrightarrow[1]{%
 \mathbin{\ooalign{$\,\xrightarrow{#1}$\cr$\xleftarrow{\hphantom{#1}}\,$}}
}

\begin{document}
\providecommand{\R}{\ensuremath \mathbb{R}}
\providecommand{\Rtri}{\ensuremath \mathbb{R}^3}
\providecommand{\Rtrisq}{\ensuremath \mathbb{R}^{3\times3}}
\providecommand{\Rsix}{\ensuremath \mathbb{R}^6}
\providecommand{\Rsixsq}{\ensuremath \mathbb{R}^{6\times6}}
\providecommand{\Rsev}{\ensuremath \mathbb{R}^7}

\providecommand{\Stri}{\ensuremath \mathbb{S}^3}
\providecommand{\N}{\ensuremath \mathbb{N}}

\providecommand{\T}{{^\top}}

\newtheorem{defn}{Definition}
\newtheorem{rem}[defn]{Remark}
\newtheorem{lem}[defn]{Lemma}
\newtheorem{prop}[defn]{Proposition}
\newtheorem{assum}[defn]{Assumption}
\newtheorem{ex}[defn]{Example}
\newtheorem{runx}{Running Example}
\newtheorem{thm}[defn]{Theorem}
\newtheorem{cor}[defn]{Corollary}
\newcommand{\regtext}[1]{\mathrm{\textnormal{#1}}}

\newcommand{\new}[1]{{\color{PineGreen}{#1}}}
\newcommand{\old}[1]{{\color{Plum}\sout{#1}}}

\providecommand{\mass}{\mathbf{M}}
\providecommand{\cori}{\mathbf{C}}
\providecommand{\grav}{\mathbf{g}}
\providecommand{\uu}{\mathbf{u_\text{ext}}}
\providecommand{\uh}{\mathbf{u}_h}
\providecommand{\ur}{\mathbf{u}_r}
\providecommand{\uj}{\boldsymbol{\tau}\jnt}
\providecommand{\Ttf}{\mathbf{T}}

\providecommand{\state}{\mathbf{x}}
\providecommand{\rstate}{\mathbf{x}_r}
\providecommand{\xpos}{\mathbf{p}}
\providecommand{\jacob}{\mathbf{J}}
\providecommand{\grasp}{\mathbf{G}}
\providecommand{\quat}{\mathbf{q}}
\providecommand{\rotv}{\boldsymbol{\omega}}
\providecommand{\rota}{\boldsymbol{\alpha}}
\providecommand{\load}{_l}
\providecommand{\robot}{_r}
\providecommand{\human}{_h}
\providecommand{\anysub}{_*}
\providecommand{\jnt}{_\theta}
\providecommand{\jntv}{\boldsymbol{\theta}}
\providecommand{\jnth}{\boldsymbol{\theta}_h}

\newcommand{\des}{^\text{d}}

\providecommand{\impD}{\mathbf{D}}
\providecommand{\impK}{\mathbf{K}}
\providecommand{\pos}{p}
\providecommand{\rot}{o}
\providecommand{\dsg}{\mathbf{x}^*}
\providecommand{\dsgpos}{\xpos^*}
\providecommand{\dsgquat}{\quat^*}
\providecommand{\dsAp}{\mathbf{A}_\pos}
\providecommand{\dsAr}{\mathbf{A}_\rot}
\providecommand{\dsf}{\mathbf{f}}
\providecommand{\dsu}{\mathbf{u}_{\text{DS}}}
\providecommand{\kq}{k_q(\quat, \dsgquat)}
\providecommand{\kqest}{k_q(\quat, \est\dsgquat)}

\providecommand{\dsgest}{\hat{\mathbf{x}}^*}
\providecommand{\pfx}{{\boldsymbol{\zeta}}}
\providecommand{\pfdyn}{{\boldsymbol{g}}}
\providecommand{\pfobf}{{\boldsymbol{h}}}
\providecommand{\pfob}{{\boldsymbol{y}}}
\providecommand{\pfdynn}{{\boldsymbol{n}}}
\providecommand{\pfobn}{{\boldsymbol{v}}}
\providecommand{\eye}{{\mathbf{I}}}
\providecommand{\tune}{{\boldsymbol{\eta}}}
\providecommand{\rand}{{\mathbf{Z}}}
\providecommand{\hfn}{{\boldsymbol{\mu}(\dsgpos_i)}}
\providecommand{\infradius}{\eta_5}
\providecommand{\ellmat}{{\mathbf{E(\boldsymbol{\theta}_h)}}}

\providecommand{\manimat}{{\mathbf{E_m}}}
\providecommand{\zerovec}{{\mathbf{0}}}
\providecommand{\var}{\boldsymbol{\sigma}^2}
\providecommand{\decay}{d}

\providecommand{\estm}{\widehat}
\providecommand{\est}[1]{\ensuremath \widehat{#1}}
\providecommand{\twonorm}[1]{\ensuremath ||#1||_2}
\providecommand{\wvp}{\ensuremath \eta_1}
\providecommand{\wap}{\ensuremath \eta_2}
\providecommand{\wvr}{\ensuremath \eta_3}
\providecommand{\war}{\ensuremath \eta_4}
\providecommand{\esterr}{\ensuremath e}

\providecommand{\ujr}{\tau_\text{lim}}
\providecommand{\ujn}{\boldsymbol{\tau}_\text{N}}

\providecommand{\ff}{\mathbf{b}}
\providecommand{\fg}{\mathbf{g}}
\providecommand{\fh}{\mathbf{h}}
\providecommand{\quatu}{\boldsymbol{v}}

\providecommand{\skewsym}{\mathbf{S}}
\providecommand{\work}{\mathbf{W}}

\newcommand{\red}[1]{{#1}}
\newcommand{\blue}[1]{{#1}}
\newcommand{\add}[1]{{#1}}

\providecommand{\Review}[3]{\textcolor{red}{{\footnotesize \marginnote{#3}}\sout{#1}\red{{#2}}}}
\providecommand{\ReviewR}[3]{\textcolor{red}{{\footnotesize \marginnote{#3}}\sout{#1}\red{{#2}}}}
\providecommand{\ReviewL}[3]{\textcolor{red}{\reversemarginpar{\footnotesize\marginnote{#3}}\sout{#1}\red{{#2}}}}
\newcommand{\cyan}[1]{{\color{cyan} #1}}
\providecommand{\ReviewNEW}[3]{\textcolor{cyan}{{\footnotesize \marginnote{#3}}\sout{#1}\cyan{{#2}}}}

\newcommand{\stkout}[1]{\ifmmode\text{\sout{\ensuremath{#1}}}\else\sout{#1}\fi}

\newcommand{\re}[1]{\begin{taggedblock}{rebute} #1 \end{taggedblock}}

\title{Constraint-Aware Intent Estimation for Dynamic Human-Robot Object Co-Manipulation}

\author{\authorblockN{Yifei Simon Shao, Tianyu Li, Shafagh Keyvanian, Pratik Chaudhari, Vijay Kumar and Nadia Figueroa}
\authorblockA{GRASP Laboratory, University of Pennsylvania, Philadelphia, PA, 19104 USA \\ {\tt\small\{yishao, tianyuli, shkey, pratikac, kumar, nadiafig\}@seas.upenn.edu}. }}

\maketitle

\begin{abstract}
Constraint-aware estimation of human intent
is essential for robots to physically collaborate and interact with humans. Further, to achieve fluid collaboration in dynamic tasks intent estimation should be achieved in real-time. In this paper, we present a framework that combines online estimation and control to facilitate robots in interpreting human intentions, and dynamically adjust their actions to assist in dynamic object co-manipulation tasks while considering both robot and human constraints. Central to our approach is the adoption of a Dynamic Systems (DS) model to represent human intent. Such a low-dimensional parameterized model, along with human manipulability and robot kinematic constraints, enables us to predict intent using a particle filter solely based on past motion data and tracking errors. For safe assistive control, we propose a variable impedance controller that adapts the robot's impedance to offer assistance based on the intent estimation confidence from the DS particle filter. We validate our framework on a challenging real-world human-robot co-manipulation task and present promising results over baselines. 
Our framework represents a significant step forward in physical human-robot collaboration (pHRC), ensuring that robot cooperative interactions with humans are both feasible and effective. \url{https://tinyurl.com/intent-capability}
\end{abstract}
\IEEEpeerreviewmaketitle
\vspace{-3.5pt}
\section{Introduction}
In recent decades, robots have become increasingly common in factories and warehouses, performing tasks that require high speeds and forces and can be dull, dirty or dangerous for humans, such as object transportation, inspection, palletization, etc. However, robots with such capabilities usually operate in confined and controlled environments, physically separated from humans with physical cages or sensor barriers to ensure safety and efficiency. On the other hand, robots currently being deployed in human-centric environments are typically lightweight, with speed and force limiting constraints hindering their ability to perform physically useful high speed or payload heavy tasks alongside humans \cite{ROB-052}.

One promising future would be for robots that have stronger physical capabilities (such as those deployed in factories and warehouses) to work intuitively alongside humans, offering physical, fluent, and safe assistance in tasks such as co-manipulation of heavy or bulky objects where neither the robot nor the human can achieve the task alone. In particular, we are interested in scenarios where the robot does not have \textit{a priori} knowledge of the task yet the human requires physical assistance to perform it. The difficulty in such tasks is discerning the human's intent. Prior work has explored using natural language \cite{brohan2023can, wu2023tidybot} or gestures \cite{gleeson2013gestures} to convey the human's guiding cues to the robot. Nevertheless, humans have a remarkable ability to co-manipulate objects with partners solely through \textit{physical interaction}, even with impaired vision or language processing. Inspired by this human ability, in this work, we seek to endow robots with the capability to estimate the human's intent solely from \textit{physical guidance} while taking into consideration kinematic and feasibility constraints of both the agents for fluid human-robot object co-manipulation tasks.

\begin{figure}[tbp] 
    \centering
    \includegraphics[ width=\linewidth]{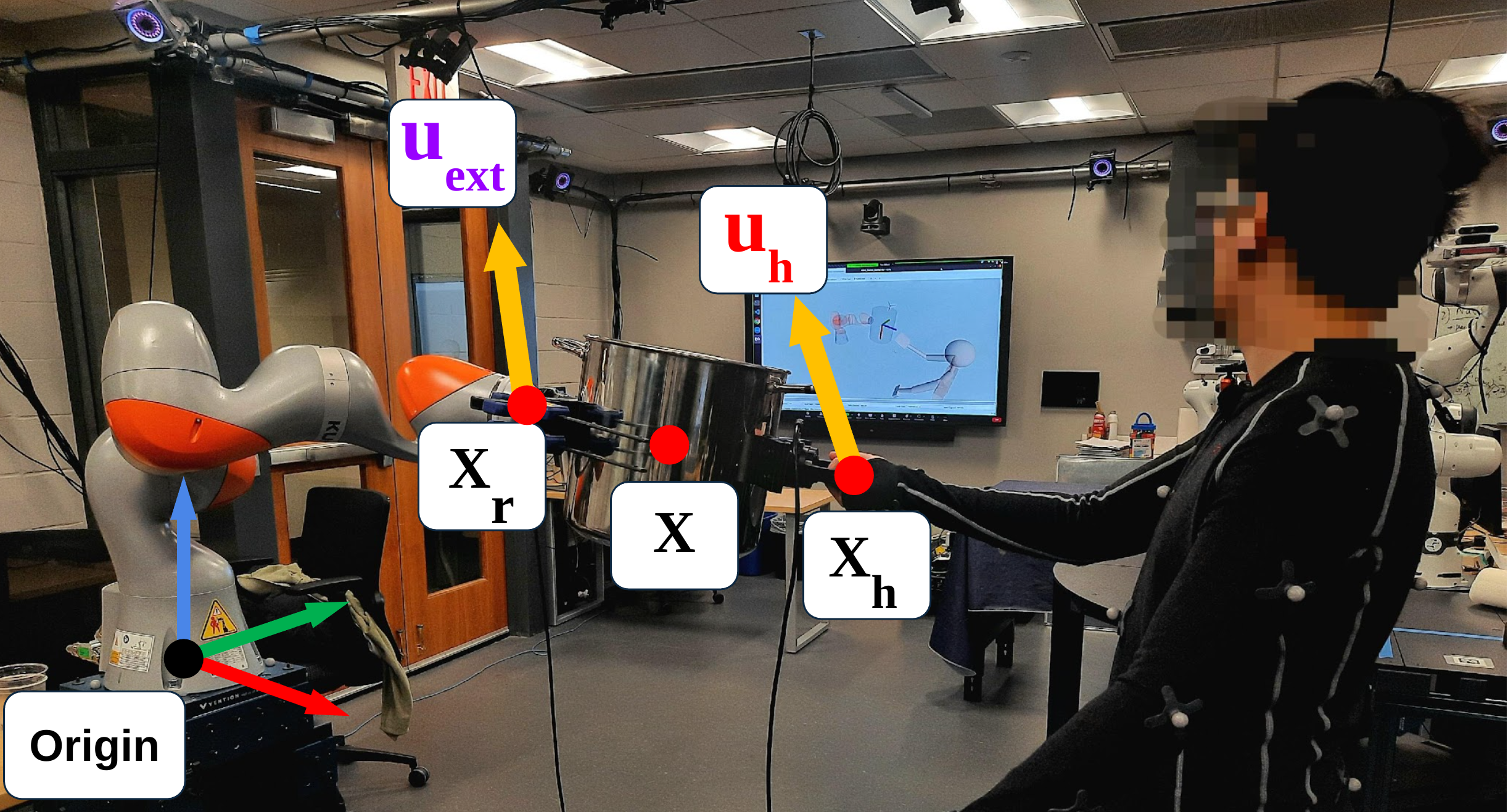}
    \vspace{-10pt}
    \caption{Our method uses particle filters to predict full 6 DoF intent and a variable impedance control scheme to assist the human, while being aware of the robot's kinematic constraints. \textit{This is achieved without any external force-torque (F/T) sensing}. Schematic representation of robot state, $\state\robot$, and force applied by human and robot in a co-carrying task. $\uu$ and $\uh$ represent the wrenches applied to the object $\state$ by the robot and the human, respectively. The red, green, and blue arrows in the inertial frame Origin correspond to the x-axis, y-axis, and z-axis.\label{fig: experiment}}
    \vspace{-15pt}
\end{figure}

Three main challenges need to be addressed to achieve this: \textbf{1) intent estimation:} how can the robot estimate the human's intent in terms of both task goal and desired motion?, \textbf{2) role adaptation:} when should the robot switch between active/passive assistance or leader/follower roles when physically interacting with the human? and \textbf{3) constraint satisfaction:} how can we ensure that kinematic and feasibility constraints for both the robot and the human are taken into consideration when performing the co-manipulation task? Next, we elaborate on each of these challenges and describe our proposed solutions; corresponding to the main contributions of this work.

\textbf{Intent estimation:} In our context, intent consists of \textit{goal estimation} and \textit{motion estimation} \cite{hoffman2023inferring}. Goal estimation includes inferring goals at different levels, such as strategic, semantic, and spatial goals, which usually represent the long-term outcome. Prior works have tackled the goal estimation problem via Bayesian methods \cite{jain2019probabilistic, iregui2021reconfigurable, losey2022learning, zanchettin2017probabilistic}, Reinforcement Learning (RL) \cite{le2021hierarchical} and supervised learning (SL) \cite{sidiropoulos2021human, choi2022preemptive} techniques. In physical human-robot collaboration (pHRC) scenarios, previous works mainly focus on predicting intended goals from a predefined set of discrete goals and adapting the behavior of the robot to switch across these set of goals \cite{khoramshahi2018human, nicolis2018human, jain2019probabilistic, ng2023takes}. 

In contrast to goal estimation, motion estimation refers to a series of spatial movements. The motion can be long horizon trajectories or immediate movements; i.e., velocities or directions of desired motion. Current techniques for motion estimation in the context of pHRC have adopted learning-based methods leveraging past motion data and/or force inputs to predict future motion \cite{ikeura1995variable,nicolis2018human, amor2014interaction, cheng2021long, al2021improving, lawitzky2012feedback, Franceschi_2023}. Nevertheless, they require extensive calibration, data collection or even transfer techniques to be used with different humans and physical guidance interfaces. Furthermore, most works focus solely on predicting translational motion and goals, with only very few works considering 6DoF (3D position and 3D orientation) co-manipulation \cite{sidiropoulos2021human, haninger2022model}. This is due to the well-known rotation-translation problem that arises when an extended object is co-manipulated. In such cases, lateral measured forces can indicate either intent to translate laterally, or intent to rotate the object in the plane, as extensively studied in \cite{Mielke_comanip_2024}. We posit that relying on force-torque sensors attached to the end-effector of the robot is not the most effective approach for intent estimation. Not only do they provide ambiguous measurements during extended object co-manipulation, but forbid the robot from reacting to human input on any part of its body or the object, making interaction unnatural. Therefore, in this work, we propose a 6DoF co-manipulation scheme for flexible and intent-aligned motions that does not require such a sensor.

Interestingly, in certain scenarios combining goal estimation and motion estimation could yield better performance  \cite{hoffman2023inferring}. Yet, such a combination has only been explored in non-co-manipulation settings where the human is moving in free space. In this work, we show that such a combined approach is also highly beneficial in co-manipulation scenarios as exposing both the motion and the goal to the algorithm, we not only gain fluid task execution but also ensure safety and feasibility.

\textbf{Role adaptation:} Here, the problem is to know when the robot should provide active assistance, versus when it should let the human take over. Such a transition might not be significant for lighter-weight robots as humans can easily overcome the robot's motion. However, correct transition is crucial in ensuring safe and effortless operation when working with robots with significant payload capacity \cite{li2015continuous}. These robots are powerful enough to cause injury when over-assisting. Existing approaches typically adjust the controller gains in discrete steps \cite{ikeura1995variable} based on the confidence of the estimator. Also, prior work like \cite{dragan2013policy} uses confidence to decide on policy arbitration for shared autonomy. Our work introduces a novel approach in which an estimation confidence measure is used to adjust the controller gains. Hence, the robot will only provide active assistance with a high-confidence estimation of intent.

\textbf{Constraint satisfaction:} The last challenge we want to address is robot and human constraints. The co-manipulation task involves a physical coupling/connection between the robot and the human via the co-manipulated object. Such a connection restricts the motion of the team depending on different capabilities of the robot and human. Specifically, both human and robot motion capability could be limited by reachability, joint limits, external collision, self-collision, etc. In this work, we consider reachability and joint limits as robot constraints, assuming there are no external obstacles and self-collision. While robot motion capability is more straightforward to model, human capability requires more careful treatments since humans have the freedom to move. Therefore, our work considers the manipulability ellipsoid as the constraint for human motion. Such manipulability ellipsoid could help generate co-manipulation motions with more comfort.

\textbf{Approach:} We propose a method that 1) estimates goal/motion simultaneously, 2) performs role adaptation, 3) while being aware of the constraints, all within dynamic co-manipulation task. We achieve these goals by adopting the Dynamical System (DS) motion policy paradigm \cite{billard2022learning}. We assume the motion of the human and the robot cam be modeled as an autonomous $\dot{x}=f(x)$ with $f(x):\mathbb{R}^N\rightarrow\mathbb{R}^N$ being the mapping from the input state $x\in\mathbb{R}^N$ to its derivative $ \dot{x}\in\mathbb{R}^N$ and converging to a target $x^*\in\mathbb{R}^N$ from anywhere in the state-space; i.e., $\lim_{t\rightarrow\infty}||x-x^*||$. As will be shown in Section \ref{sec:approach}, such a low-dimensional parameterized representation of desired motion can be used to estimate both goal $x^*$ and motion $\dot{x}$ online in real-time and is amenable and compatible with compliant control strategies and adaptation mechanisms for physical human-robot interaction (pHRI) tasks \cite{khoramshahi2019dynamical, khoramshahi2018human}. Hence, the major \textbf{contributions} of this work are as follows:
\begin{itemize}
    \item We predict human intent, including goal and motion, in co-manipulation tasks by estimate the parameters of a 6DoF DS motion policy using particle filters driven by velocity tracking errors, without force-torque sensing.  
    \item We propose a confidence-based variable impedance control scheme suited to track the estimated 6DoF DS motion policy. By estimating the confidence of our particle filter predictions our method fluently decides when to offer active or passive assistance without a force/torque sensor.
    \item We ensure goal/motion feasibility during intent estimation by trimmimg and reshaping particles using robot kinematic constraints and human manipulability measures.
    \item We validate our proposed constraint-aware intent estimation approach on a challenging co-manipulation task with real hardware experiments, showcasing improved performance over state-of-the-art techniques. 
\end{itemize}

To the best of the authors' knowledge, this is the first time an approach is proposed that i) combines goal and motion intent estimation, and ii) considers a quaternion-based 6 DoF intent estimation for the human-robot co-manipulation tasks.

\textbf{Paper Organization:} We begin in Section \ref{sec:rw} by summarizing relevant works that tackle intent estimation and motion generation for co-manipulation tasks. In Section \ref{sec:problem-formulation} we introduce the mathematical problem formulation and assumptions taken in this work. Our proposed approach leveraging DS for motion/goal representation, constraint-aware particle filters for estimation and confidence-based variable impedance control for compliant and adaptive control is introduced in Section \ref{sec:approach}. Finally, in Section \ref{sec:experiments} we validate our approach on a real hardware co-manipulation tasks and evaluate it against baseline techniques showcasing improved task performance and minimization of human effort during collaboration.

\section{Related Work}
\label{sec:rw}

Several prior works have proposed approaches for human-robot co-manipulation, which involves tackling multiple challenges listed in the previous section. 

Among these works, some consider exploiting learning-based methods, especially for the intent estimation component. For example, \cite{haninger2022model} uses Gaussian Process (GP) to perform intent recognition, with Model Predictive Control (MPC) being the motion generator and admittance control as the low-level control. \cite{nicolis2018human} uses Recurrent Neural Network (RNN) to predict and classify intent trajectory and uses admittance control for the low-level motion. \cite{al2021improving} predicts robot desired velocity from human force input with Weighted Random Forest (WRF). \cite{lawitzky2012feedback} combines the learning-based method and planning-based method to exploit the complementary strengths from two sides. \cite{rozo2014learning}\cite{rozo2016learning} learns the task from the trajectory of position, forces, and stiffness from human demonstrations with TP-GMM and reproduces motions. Learning-based approaches usually require offline effort to collect data that could be difficult to generalize to unseen tasks. We consider only having access to limited prior knowledge or data about the potential co-manipulation tasks. \cite{Franceschi_2023} uses RNN with a final fully connected layer for transfer learning on different human subjects to alleviate such a problem. However, extra effort is still required to collect data specific to different tasks and subjects. Also, out-of-distribution scenarios could cause concern as the robot could injure humans through physical interaction.

There are model-based approaches with online adaptation schemes. \cite{al1997arm} designs a bio-inspired two-level reflexive motion control, using a high-level force threshold to trigger the lower-level compliant control. \cite{ficuciello2015variable} proposes a variable impedance controller that varies mass and damping properties of the robot based on human input velocities with consideration of stability. \cite{erden2010human} proposes to use control effort to infer human intent without a force sensor. \cite{khoramshahi2019dynamical} uses the same DS representation as we do to represent tasks but uses how well the velocity aligns with several predefined DSs for selecting which goal to help the human towards. In \cite{khoramshahi2020dynamical}, they further added an admittance controller and a power pass filter for the robot to help the human only when the human force input is small. However, these methods still require predefined goal poses and require a force/torque sensor to work well. \cite{khoramshahi2018human} uses human demonstration to estimate either a limit cycle or a goal-oriented DS from a single demonstration. However, they do not consider the case where the human holds a heavy object, nor do they vary the impedance gains. All model-based methods mentioned above consider tasks in translational space, while our work considers the full 6D pose.

Many of the works mentioned here \cite{haninger2022model, nicolis2018human, lawitzky2012feedback} assume a set of discrete goals. Another line of work forgoes the assumption of predefined goals and investigates finding them with physical interaction. Often, some form of variable gain control is required for different behaviors depending on the confidence.
\cite{thobbi2011using} estimates future human motion using EKF and weighs the proactive controller and reactive controller based on the confidence of the estimation.  
\cite{duchaine2007general} estimates human intent using the derivative of force and changes the damping gain of the robot to allow quick acceleration or deceleration.
\cite{stouraitis2020online}\cite{stouraitis2018dyadic} uses hybrid optimization to solve for a collaborative 14 DoF two-handed manipulation tasks, with a simple PD controller for the human intent prediction.
\cite{li2019differential} \cite{takagi2020flexible} builds an intent estimator together with the robot controller gains. They show that even without an accurate estimation of the intent, the robot can still achieve good performance with guarantees. However, only one-dimensional case is considered.

\vspace{-3.5pt}
\section{Problem Formulation}
\label{sec:problem-formulation}
 \begin{assum}
 We assume a rigid object, with known geometry and CoM (center of mass) pose $\state=[\mathbf{p}, \quat] \in \Rtri\times\Stri$ is rigidly attached to an $n$-DoF robot at a known contact pose $\state\robot=[\mathbf{p}\robot, \quat\robot] \in \Rtri\times\Stri$ and is to be co-manipulated with a human partner as depicted in Fig. \ref{fig: experiment}. We assume the contact pose of the human $\state\human=[\mathbf{p}\human, \quat\human] \in \Rtri\times\Stri$ is also known, with $\mathbf{p}\anysub\in\mathbb{R}^3$ being positions and $\quat\anysub = [s\anysub, \quatu\anysub] \in \Stri \subset \R^4$ unit quaternions representing the orientations expressed in world frame coordinates for all known poses.
 \label{assum: known_poses}
\end{assum}

\textbf{Object Dynamics:} We begin by introducing the dynamics model of the object being co-manipulated by an $n$-DoF robot and a human, which can be expressed as follows,
\begin{align}
    \mass\load \ddot{\state}+\grav\load=\grasp^\top_r\textcolor{violet}{\uu} + \grasp^\top_h\textcolor{red}{\uh}\label{eq:obj_dyn},
\end{align}
\noindent 
where $\mass\load \in \Rsixsq$ is the object mass/inertia matrix, $\grav\load\in\Rsix$ is the gravitational force of the object and $\ddot{\state}\in\Rsix$ corresponds to the acceleration of the object's CoM. Further, $\uu, \uh \in \Rsix$ are the wrenches applied on the object by the robot and the human, at known contact poses $\state\robot$ and $\state\human$ respectively, as in Fig. \ref{fig: experiment}. Finally,  $\grasp\anysub^\top \in \Rsixsq $ represent the known full-rank grasp matrices transforming the wrenches applied at the known contact poses to the CoM of the object.
 \begin{assum}
    We assume there is an \textbf{accurate} model of $\mass\load$ and consequently $\grav\load$ and since the object is rigidly attached to the robot, the grasping matrices $\grasp\robot$ and $\grasp\human$ are constant.
     \label{assum: rigid_attach}
\end{assum}

\begin{assum}
    We assume that we \textbf{do not} have a \textbf{model of the human dynamics} nor access to accurate measurements of the human wrench $\textcolor{red}{\uh}$ exerted on the contact pose $\state\human$.
     \label{assum: human_dynamics}
\end{assum}

\textbf{Robot Dynamics:} The robot rigidly attached to the object is an $n$-DoF torque-controlled manipulator mounted on a stationary base with rigid-body dynamics derived by the Euler-Lagrange equations and defined as \cite{spong2nd}:
\begin{align}
    \mass\jnt(\jntv)\ddot{\jntv}+\cori\jnt(\jntv, \dot{\jntv}) \dot{\jntv} +\grav\jnt(\jntv)=\uj \textcolor{violet}{-}\jacob(\jntv)^\top\textcolor{violet}{\uu}, \label{eq: robot_dyn_joint}
\end{align}
where $\jntv,\dot{\jntv}, \ddot{\jntv} \in \mathbb{R}^n$ denote the joint positions, velocities, and accelerations. $\mass\jnt(\jntv)\in \mathbb{R}^{n\times n}$, $\cori\jnt(\jntv, \dot{\jntv})\in \mathbb{R}^{n\times n}$, $\grav\jnt(\jntv)\in \mathbb{R}^{n}$ denote the inertia matrix, Coriolis matrix and gravity respectively. $\uj\in \mathbb{R}^n$ denotes the robot control input/command. Notably, \textcolor{violet}{$-\uu$} denotes the reaction wrench experienced by the robot from the object co-manipulation interaction at contact pose $\state\robot$, exhibiting an opposite sign from Eq. \eqref{eq:obj_dyn}. This external wrench is transformed to an external torque vector via the manipulator Jacobian $\jacob(\jntv)\in\mathbb{R}^{6\times n}$. In order to make a connection between the robot rigid-body dynamics introduced in Eq. \eqref{eq: robot_dyn_joint} and the object dynamics introduced in Eq. \eqref{eq:obj_dyn}, we can express the dynamics model of the robot in operational space as follows \cite{1087068}, 
\begin{align}
\mass\robot(\jntv) \ddot{\state}\robot+\cori\robot(\jntv, \dot\jntv) \dot{\state}\robot+\grav\robot(\jntv)={\color{Green}\ur} - \textcolor{violet}{\uu} \label{eq: robot_dyn_cart},
\end{align}
where $\mass\robot(\jntv)\in \mathbb{R}^{6\times 6}, \cori\robot(\jntv, \dot\jntv)\in \mathbb{R}^{6\times 6}, \grav\robot(\jntv)\in \mathbb{R}^{6}$ now correspond to the apparent mass, Coriolis, and gravity of the manipulator in operational space coordinates; with $\dot{\state}\robot,\ddot{\state}\robot\in\mathbb{R}^6$ being the velocity and acceleration of the robot's end-effector, respectively. \textcolor{violet}{$\uu$} is the same external wrench as in Eq.~\eqref{eq:obj_dyn} and \eqref{eq: robot_dyn_joint} and $\color{Green}\ur$ corresponds to the operational space robot input which is computed as ${\color{Green}\ur} = \jacob(\jntv)\T^\dag\uj$ stemming from $\uj = \jacob(\jntv)^\top {\color{Green}\ur}$, with $(\cdot)^\dag$ denoting the pseudo-inverse of a matrix. This mapping between control wrench and torque ${\color{Green}\ur} \xleftrightarrow{} \uj$ is only bijective when $n=6$ and $\jacob(\jntv)$ is full rank. In our case, since we are working with a redundant $n=7$ DoF manipulator, the solution $\uj$ is not unique, therefore we have the possibility to choose $\uj$ to satisfy additional requirements such as singularity avoidance, joint limit avoidance and any other relevant constraints as shown in Section \ref{sec:robot-control}.

\textbf{Robot-Object Dynamics:} Without loss of generality, since $\grasp\human$ is invertible, we can assume the human acts on the CoM of the object and express the unknown human wrench, \textcolor{red}{$\uh$}, in CoM coordinates as
$\textcolor{red}{\uh'}=\grasp\human\T\textcolor{red}{\uh}$. Thus, by plugging Eq.~\eqref{eq: robot_dyn_cart} into Eq.~\eqref{eq:obj_dyn} we can define the dynamics model of the robot-object pair in the object CoM frame as follows,
\begin{align}   
\label{eq:combined_dyn_simpl}
    \mass \ddot\state+\cori\dot\state + \grav = \grasp\robot^\top{\color{Green}\ur} + \textcolor{red}{\uh'},
\end{align} 
with $\mass = \mass\load + \grasp\robot^\top \mass\robot \in \mathbb{R}^{6\times 6}$ and $\cori = \grasp\robot^\top \cori\robot \in \mathbb{R}^{6\times 6}$ being the mass/inertia and Coriolis matrices of the combined robot-object system in CoM object coordinates. $\grav = \grasp\robot^\top\grav\robot+\grav\load \in\mathbb{R}^6$ is the vector of combined gravitational forces from robot and load acting on the object's CoM.

\textbf{Goal} Given the robot-object dynamics defined in Eq.~\eqref{eq:combined_dyn_simpl}, the goal of this paper is to design the control input ${\color{Green}\ur}$ (consequently $\uj$) to minimize $\textcolor{red}{\uh'}$, without having an accurate measurement of $\textcolor{red}{\uh}$ while being aware of the feasibility constraints and capability of both the human and the robot.

\begin{figure}[t]
    \centering
    \includegraphics[trim={3cm 25.5cm 8cm 1cm},clip, width=\linewidth]{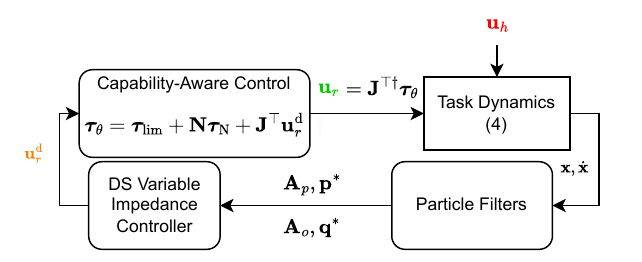}
    \caption{Control diagram of our proposed approach. We estimate dynamics matrices $\est{\dsAp}, \est{\dsAr}$ and attractors $\est\dsgpos, \est\dsgquat$ online, these estimates along with their confidence $c_\pos, c_\rot$ are sent to the DS Variable Impedance Controller to generate desired force $\textcolor{orange}{\ur\des}$. Lastly, a torque $\uj$ satisfying the constraints is computed to send to the robot. Eq. \eqref{eq:combined_dyn_simpl} refers to the combined dynamics in the Cartesian space.}
    \label{fig: ctrl_diagram}
    \vspace{-15pt}
\end{figure}

\section{Approach}
\label{sec:approach}

We aim at solving two sub-problems with an approach that tightly couples intent estimation and control (see Fig. \ref{fig: ctrl_diagram}):
\begin{enumerate}
\item Use estimation techniques for human goal and motion intent, taking into account human and robot constraints. 
\item When the intent estimation is confident, drive the robot to minimize the human effort. Otherwise, make the robot fully passive; i.e., only perform gravity compensation.
\end{enumerate}
To solve solving \textit{sub-problem 1)} we begin by modeling the \textbf{human intent} as a goal-oriented autonomous dynamical system \textbf{(DS) (Section \ref{sec:intent-ds})}. We assume that the parameters of such DS are unknown, hence in Section \ref{sec:pf-ds} we propose to use \textbf{particle filters} to estimate them. We then formulate how we compute confidence as a function of tracking error for dynamic gain adjustment in Section \ref{sec:confidence}. To take into account human's capability based on their posture we use manipulability ellipsoids to locally reshape the \textbf{variable noise} of the particles in Section \ref{sec:manip-noise}, directly considering human capabilities in our estimates. To account for the capability constraints of the team during motion and at the goal, we perform \textbf{particle trimming} for infeasible intent particles, as detailed in Section \ref{sec:pf-trimming}. 

To solve \textit{sub-problem 2)} while avoiding the robot from applying counterproductive wrenches, we propose a \textbf{variable impedance control} scheme to ensure the robot only applies strong force when the confidence of the intent estimate is high, as outlined in Section \ref{sec:conf-vic}. Lastly, to account for the capability constraints of the robot during motion, we use a \textbf{joint-limits-aware controller} to realize the desired control input, described in Section \ref{sec:robot-control}. Fig. \ref{fig: ctrl_diagram} illustrates the control diagram with each piece realized. 
\subsection{Modeling Human Intent as a Dynamical System}
\label{sec:intent-ds}
Human intent is difficult to model, and we do not attempt to model it accurately. 
Instead, we try to find a representation of intent that i) combines motion and goal estimation, ii) is approximately accurate, iii) enjoys simple parameterization and iv) amenable for compliant control. As shown in many prior works \cite{billard2022learning,KRONANDER201552,khoramshahi2018human,khoramshahi2019dynamical,khoramshahi2020dynamical} representing desired motion of a robot (or human) via DS meets all of these requirements, while enjoying theoretical guarantees such as stability, convergence and robustness to perturbations.
\begin{assum}
\label{ass:ds_assumption}
    Human intent when co-manipulating an object with a robot can be modeled approximately as the human desiring the \textbf{object's CoM pose} to follow a goal-oriented first-order Cartesian and Rotational autonomous DS, as,
\begin{align}
        \dot{\xpos} &=\dsf_\pos(\xpos;\dsAp, \dsgpos) =  \dsAp(\xpos - \dsgpos),\label{eq:ds_pos_dot}\\
        \rotv &=\dsf_\rot(\quat;\dsAr, \dsgquat)=  \dsAr \kq \log (\quat \otimes \bar\dsgquat), \label{eq:ds_rot_dot}
\end{align}
    \noindent where $\rotv\in \Rtri$ the time derivative of $\quat$, $\dsg = [\dsgpos, \dsgquat]$ the position and quaternion \textbf{attractors}, $\dsAp, \dsAr \in \Rtrisq$ \textbf{diagonal} $\prec 0$ \textbf{dynamics matrices} and $\kq= \frac{||\text{vec}(\quat \otimes \bar{\quat}^*)||}{\arccos(\text{scalar}(\quat \otimes \bar{\quat}^{*}))}$.
\end{assum}
It is well-known that the Cartesian DS defined in Eq.~\eqref{eq:ds_pos_dot} is globally asymptotically stable (GAS) if $\dsAp\prec 0$ \cite{billard2022learning}. As stated in the following theorem, for the rotational DS presented in Eq.~\eqref{eq:ds_rot_dot}, we prove that as long as the conditions defined in Eq.~\eqref{eq:stability_cond_quat} hold it is also GAS towards $\dsgquat$. Note that we provide the mathematical preliminaries of quaternion arithmetic and operations used in this work in Appendix \ref{app: quat_prelims}.
\begin{thm}
\label{thm:orient_DS}
The DS defined in \eqref{eq:ds_rot_dot} is globally asymptotically stable (GAS) at the attractor $\quat^{*}$, i.e. 
\begin{equation}
\lim_{t\rightarrow \infty}||\log( \quat^{*} \otimes \bar{\quat}(t))||=0
\end{equation}
if the following conditions hold,
\begin{equation}
\label{eq:stability_cond_quat}
\begin{cases}
\mathbf{A}_{o} = \mathbf{A}_{o}^T \prec 0\\
k_q(\quat,\quat^{*}) = \frac{||\text{vec}(\quat \otimes \bar{\quat}^*)||}{\arccos(\text{scalar}(\quat \otimes \bar{\quat}^{*}))}
\end{cases}
\end{equation}
\noindent \textbf{Proof:} See Appendix \ref{sec:linear_quatDS}. $\hfill \blacksquare$
\end{thm}

\textbf{Main idea:} Following Assumption \ref{ass:ds_assumption} both dynamics matrices are diagonal and $\dsAp, \dsAr \prec 0$. As will be shown in the next sub-section, for any position and quaternion goal estimate $\dsg = [\dsgpos, \dsgquat]$, as long as we guarantee our estimates of $\dsAp, \dsAr$ to be negative definite, the estimated DS for Cartesian and Rotational motion intent will be \textbf{GAS by construction}.

\subsection{Capability-Aware Intent Estimation using Particle Filters}
We propose to use particle filters and Bayesian updates to estimate the probability distribution of the attractors $\dsg = [\dsgpos, \dsgquat]$ and the dynamics matrices $\dsAp, \dsAr \in \Rtrisq$ from past motion data. 
Since we need to consider explicit constraints for current motion and the desired goal, using particle filters allows us to treat each estimated DS (particle) individually.

\subsubsection{Particle Filter DS Parameter Estimation}
\label{sec:pf-ds}
We use two independent particle filters for estimation, one for Cartesian DS, with states $\pfx_\pos = [\dsAp, \dsgpos]$, and the other for Rotational DS, with states $\pfx_\rot =[\dsAr, \dsgquat]$. 
The particle filter uses the dynamics model $\dot\pfx = \pfdyn(\pfx, \pfdynn)$ and observation model $\pfob = \pfobf(\pfx)$ to estimate the probability distribution of the state $\pfx$ given the observation $\pfob$. $\pfdynn$ is the process noise. Note this is an unconventional use of particle filters. Usually the filtered state denotes the robot position and changes as a function of time but here we are estimating intent in terms of DS parameters, which can be stationary even with robot motion.

We employ a zero dynamics model $\dot\pfx_\pos = \pfdynn_\pos, \dot\pfx_\rot = \pfdynn_\rot$ for the human intent.
The observations are $\pfob_\pos = [\dot\xpos, \ddot\xpos], \pfob_\rot = [\rotv, \rota] \in\mathbb{R}^6$, where $\rota = \dot{\rotv} \in\mathbb{R}^3$ is the time derivative of $\rotv$. 
We intentionally do not use force as an observation to enable the approach to be used in a wider range of scenarios, and further alleviate the shortcomings of using F/T sensors in such safety-critical estimation tasks. Using Assumption \ref{ass:ds_assumption}, the observation models for velocities are naturally,
\begin{align} %
    \est{\dot\xpos} &= \est\dsAp(\xpos - \est\dsgpos)\label{eq: Apest},\\
    \est{\rotv} &= \est\dsAr \kqest \log (\quat \otimes \est{\bar\dsgquat}).\label{eq: Arest}
\end{align}
To gain independent control over individual components of the estimation, we take the derivative of Eq. \eqref{eq: Apest} and Eq. \eqref{eq: Arest} to estimate
\begin{align}
    \est{\ddot\xpos} &= \est\dsAp \dot\xpos,\label{eq: Apest2}\\
    \est{\rota} &= \est\dsAr \frac{1}{2}\begin{bmatrix}
        \est{\quatu^*}\quatu^T + (-\est{s^*}\mathbf{I} + \skewsym(\est{\quatu^*})) (s\mathbf{I} - \skewsym(\quatu))\\ 
      \end{bmatrix} \rotv\label{eq: Aoest2}, 
\end{align}
\noindent where $\eye$ is the identity matrix of appropriate size, $\dot\xpos, \rotv$ are now taken as given rather than estimated, and $\skewsym(\cdot)$ the skew-symmetric transformation of the vector. 

The weights for a particle $i$ are estimated as follows, 
\begin{align}
    w_\pos^i &= \exp(-\wvp\twonorm{\dot\xpos - \est{\dot\xpos}_i}^2 - \wap\twonorm{\ddot\xpos - \est{\ddot\xpos}_i}^2),\\
    w_\rot^i &= \exp(-\wvr\twonorm{\rotv - \est{\rotv}_i}^2 - \war\twonorm{\rota - \est{\rota}_i}^2).
    \label{eq:particle_weights}
\end{align}
\noindent where $\eta_*$ are tunable weights, which we will use throughout the paper. The particles are then reweighed according to the weights for the next iteration. To obtain intent states estimates, $\est{\pfx_\pos},\est{\pfx_\rot}$, we use the weighted average of the particles' states.

\begin{lem}
\label{lemma:stab_pf}
 The estimated parameters for DS $\dsf_\pos$ and $\dsf_\rot$ defined in Eq. \eqref{eq:ds_pos_dot} and \eqref{eq:ds_rot_dot} with states $\est{\pfx_\pos}=[\dsAr,\dsgpos], \est{\pfx_\rot}=[\dsAp,\quat^{*}]$ estimated by the particle filters with observation models Eq.~\eqref{eq: Apest} and ~\eqref{eq: Arest} are GAS at their attractors $\dsgpos$ and $\quat^{*}$.
\end{lem}
\vspace{-5pt}
\noindent\textbf{Proof:} See Appendix \ref{app: proofpf}. $\hfill \blacksquare$
\vspace{5pt}
\subsubsection{Confidence of Parameter Estimation}
\label{sec:confidence}
To help the human only when the robot is confident, we define a confidence measure $c$ for our DS parameter estimates which will also be used to vary an impedance gain to track the desired DS estimated in Section \ref{sec:conf-vic}. Several definitions of confidence can be used for a particle filter, such as aggregating the probability mass near the goal that is currently being estimated. However, we find that any human force input would change the estimated intent and the estimator will remain confident, making intent change difficult. As an alternative, we use velocity tracking errors, which immediately increases if the human exerts force on the robot. We find empirically this allows the robot to appropriately reduce confidence when subject to external forces. Furthermore, since we vary the gains as velocity tracking error increases, when the robot hits an object, either with an obstacle/human or internally such as a joint limit, it will automatically decrease gains and adjust the intent estimate, limiting the force exerted on the environment.

Therefore, we first compute velocity tracking error, separately for linear and rotational dimensions as follows,
\begin{align}
\label{eq:track_errors}
\esterr_{p}(t) = & \twonorm{\est{\dot\xpos} - \dot\xpos},\\
\esterr_{o}(t) = & \twonorm{\est{\rotv} - \rotv}.
\end{align}
To reduce influence from noise, and to properly react to the errors over time from a consistent force input, we integrate the tracking error of the filters with constant ascent rates $\decay_p, \decay_o \in \mathbb R_{> 0}$, to obtain

\begin{align}
    c_{p}(t) & = \int_{-T}^0 (\decay_{p}- e_{p}(s)) \dd{s},\label{eq:conf_p} \\ 
    c_{o}(t) & = \int_{-T}^0 (\decay_{o} - e_{o}(s)) \dd{s}, \label{eq:conf_o}
\end{align}

\noindent where $c_p, c_o \in \mathbb R_{\ge 0}$ are clipped to $ [0,1]$, and $[-T,0]$ is the time interval. Effectively, low confidence allows the human to lead when the tracking error is consistently high due to external input. 

\subsubsection{Human Manipulability Guided Process Noise} 
\label{sec:manip-noise}
The noise terms $\pfdynn_\pos, \pfdynn_\rot$ each represent the uncertainty of the attractor and the dynamics matrices in their filter, respectively. We would like the noise to be dynamically guided by two factors: the confidence, and the human posture. When the confidence of the estimate is low, we want the filter to increase the noise to allow intent particles in more state space. When the opposite is true, we do not want noise to change the estimated intent. For human posture, previous work has shown the directional intention of motion is known to be aligned with the primary axis of the manipulability ellipsoid of the human hand \cite{cos2011influence, jacquier2012manipulability, jaquier2020analysis}; hence, we shape the noise of the Cartesian DS based on the manipulability of the human.

\begin{figure}[tbp]
    \centering
    \includegraphics[trim={0 0.5cm 0 0.5cm},clip, width=0.98\linewidth]{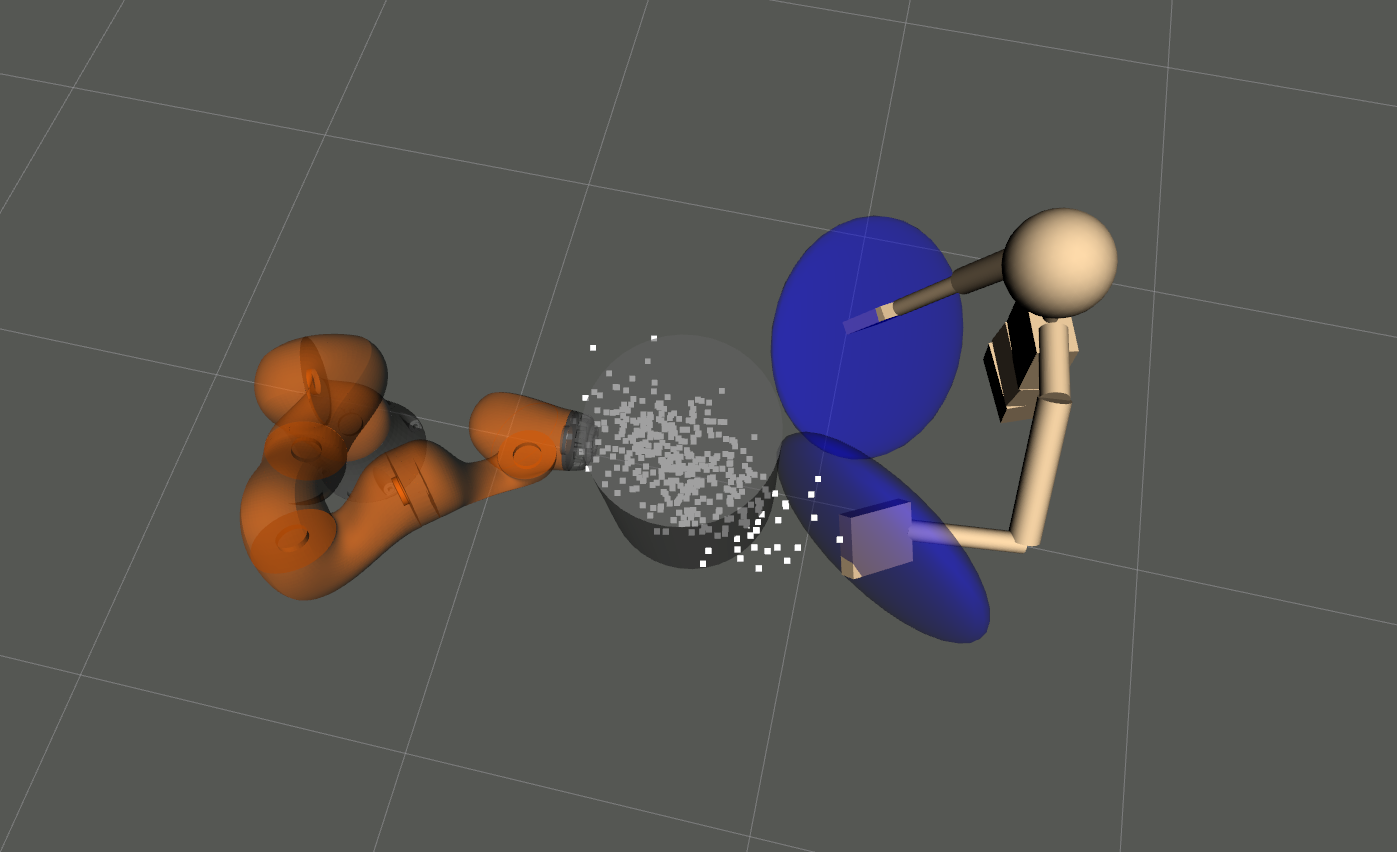}
    \caption{A demonstration of the human and robot co-carrying. Motion Capture system is employed to detect human kinematics. The manipulability ellipsoids of both human hands are depicted in blue. The ellipsoid associated with the hand closer to the pot is utilized to locally reshape the particles' noise, thereby influencing the goal distribution. Particles are visualized as white dots}
    \label{fig: rviz}
    \vspace{-13pt}
\end{figure}

Inspired by the method proposed in \cite{KRONANDER201552}, 
we reshape the variable noise by premultiplying the manipulability ellipsoid, $\ellmat = \jacob(\jnth)^\top\jacob(\jnth) \in\mathbb{S}_{++}^{3}$, extracted from human posture,
where $\jnth$ represents the joint angles of the kinematic chain of the hand closer to the object and $\jacob(\jnth)\in\mathbb{R}^{3\times n}$ the Jacobian of the human kinematic chain of the corresponding hand starting from the pelvis joint. This influence is shown in Fig. \ref{fig: rviz}.
To only locally modulate the noise since farther intended goal should not be affected by the manipulability index, we define the \textbf{local} manipulability ellipsoid matrix as
\begin{align}\label{eq:Mlocal}
    & \manimat = (1-\hfn)\eye + \hfn \ellmat,
\end{align} 
where function $\hfn$ is used to control the radius of influence, defined as 
\begin{align*}
    & \hfn = \exp \left( -\infradius \lVert \dsgpos_i -\xpos\human \rVert _2^2 \right).
\end{align*}  
with $\xpos\human$ being the position of the human hand tracked by motion capture system. Lastly, using confidence to adjust the size of the noise, we can then write the noise of the particle filters states $\pfx_\pos = [\dsAp, \dsgpos]$ and $\pfx_\rot =[\dsAr, \dsgquat]$ as
\begin{align}
    \pfdynn_\pos &= (1-c_\pos(t)) [\eye \rand(\tune_6), \rand_m],\\
    \pfdynn_\rot &= (1-c_\rot(t)) [\eye \rand(\tune_7), \rand(\tune_8)]\label{eq: rot_noise},
\end{align}
\noindent
where $\rand(\tune_*)\in \Rtri \sim \mathcal{N}(\zerovec, \eye\tune_*), \rand_m \in \Rtri \sim \mathcal{N}(\zerovec, \manimat)$ random vectors that are multivariate Gaussian, with $\tune_*\in \Rtri$ tunable noises, and $\rand(\tune_8)$ represents the noise on $\rotv$. This noise can be added on a quaternion state using Eq. \eqref{eq:quat_int} in Appendix \ref{app: quat_prelims}. 
\subsubsection{Constraint-aware Particle Trimming: Motion and Goal Feasibility}
\label{sec:pf-trimming}
Our proposed method uses the DS framework for combined motion/goal estimation, corresponding to each particle proposing motion $\est{\dot\xpos}$, and goal $\est\dsgpos$ (and their rotational counterpart angular motion $\est{\rotv}$ and goal $\est\dsgquat$). This representation allows us to analytically trim particles based on various constraints on both.
In this work, we find manipulability ellipsoid used previously sufficiently satify the motion and goal constraint for the human. Thus, we mainly consider the motion and goal constraint of the robot by using a particle trimming scheme to bring particles that are infeasible either for the \textit{immediate motion} or the \textit{goal} to zero weight. That is, we establish one-to-one correspondence for the particles of the two filters and check if motion $[\est{\dot\xpos},\est{\rotv}]$ and goal state $[\est\dsgpos, \est\dsgquat]$ is feasible using an inverse kinematics (IK) solver. If not, we set $w_\pos^i , w_\rot^i$ in Eq.~\eqref{eq:particle_weights} to zero.
For computational efficiency, we employ a recently proposed GPU-accelerated IK solver Curobo \cite{sundaralingam2023curobo} to achieve this. 

\begin{lem}
    The estimated DS $\dsf_\pos$ and $\dsf_\rot$ defined in Eq. \eqref{eq:ds_pos_dot} and \eqref{eq:ds_rot_dot} with states $\est{\pfx_\pos}, \est{\pfx_\rot}$ estimated by the particle filters with observation models Eq.~\eqref{eq: Apest} and ~\eqref{eq: Arest} are task space feasible.
 \end{lem}
 \vspace{-5pt}
\noindent\textbf{Proof:} As a result of particle trimming by Curobo for kinematically infeasible goals, the attractor is feasible in task space. Furthermore, since the goal is globally asymptotically stable and joint limits not reached thanks to the capability-aware torque controller defined in Section \ref{sec:robot-control}, the estimated DS are guaranteed task space feasible, and joint space feasible when the Jacobian of the manipulator $\jacob$ is full rank.
$\hfill \blacksquare$
\vspace{-3.5pt}
\subsection{Confidence-based Variable Impedance Control}
\label{sec:conf-vic}
Early works in co-manipulation first explored the idea of optimizing impedance gains by minimizing a cost function to achieve an optimal damping factor along a trajectory leveraging force sensors \cite{ikeura2002optimal}. However, more recent work \cite{fumery2021biomechanical} has varied conclusions about human-human collaboration when carrying a heavier object.
Therefore, for better theoretical understanding and simplicity, we vary the impedance gains \textbf{linearly} between bounds.

Since we assume the human intent is a DS, we propose to track the estimated desired velocities with the following variable damping impedance control law inspired by \cite{passiveDS,khoramshahi2020dynamical},
\begin{align}\label{eq:u_ds}
    \dsu &= -\left[\begin{array}{cc}
        \impD_{p}(c_{p}(t)) & 0 \\
        0 & \impD_{o}(c_{o}(t))
        \end{array}\right]\left[\begin{array}{l}
        \dot{\xpos}-\est{\dot\xpos} \\
        \rotv-\est{\rotv} 
        \end{array}\right],
\end{align}
with variable damping matrices defined as, 
\begin{align}\label{eq:Damp_u}
    \impD_{p}(c_{p}(t)) &= c_{p}(t)\mathbf{\Lambda}_p \succeq 0, ~~ \mathbf{\Lambda}_p = \begin{bmatrix}
    d_{p}^1 &  & 0 \\
    0 & d_{p}^2 &  0\\
    0 & 0 & d_{p}^3 \\
    \end{bmatrix}\\
\impD_{o}(c_{o}(t)) & = c_{o}(t)\mathbf{\Lambda}_o\succeq 0, ~~ \mathbf{\Lambda}_o=\begin{bmatrix}
    d_{o}^1 &  & 0 \\
    0 & d_{o}^2 &  0\\
    0 & 0 & d_{o}^3 \\
    \end{bmatrix}
\end{align}
and $\est{\dot\xpos} = \dsf_\pos(\xpos; \est{\pfx_\pos}) , \est{\rotv} = \dsf_\rot(\quat;\est{\pfx_\rot})$ are defined by the estimated DS from Eq. \eqref{eq: Apest} and \eqref{eq: Arest}. Intuitively, Eq. \eqref{eq:u_ds} is a negative velocity feedback controller that allows a robot to track the desired DS while being passive and compliant to external forces imposed by the human, similar to \cite{KRONANDER201552,khoramshahi2020dynamical}. However, contrary to such prior works, where the damping matrices are either assumed to be constant \cite{khoramshahi2020dynamical} or constructed such that state-varying terms only affect the velocity feedback term as in \cite{passiveDS}, our \textbf{damping matrices vary wrt. the time-varying confidence estimates} defined in Eq. \eqref{eq:conf_p} and \eqref{eq:conf_o}. Note that, while linearly varying wrt. corresponding velocity tracking errors, $\impD_{p}(c_{p}(t)), \impD_{o}(c_{o}(t))$ will always be positive semi-definite as $c_{p}(t), c_{o}(t) \in [0,1]$ and $d_{p}^i, d_{o}^i \in \mathbb{R}_+ \forall i=1,2,3$.

Additionally, we compensate for the gravity terms of both the robot and the object, and neglect object Coriolis effect as it's small. The \textbf{desired} control input wrench for the robot that tracks the estimated DS when the confidence is high, and compensates for gravity otherwise is, 
\begin{align}
    \textcolor{orange}{\ur\des} = \grav\robot +\grasp\robot^-\T\left(\grav\load + \dsu\right).
\end{align}

\textbf{Closed-Loop Dynamics of Controlled System}
We now analyze the closed-loop dynamics of the combined object-robot systems defined in Eq. \eqref{eq:combined_dyn_simpl} under the confidence-based variable impedance control law defined in Eq. \eqref{eq:u_ds}. Assuming perfect control tracking and gravity compensation; i.e.,  $\textcolor{green}{\ur}=\textcolor{orange}{\ur\des}=\grav\robot +\grasp\robot^-\T\left(\grav\load + \dsu\right)$ and $\grav = \grasp\robot\T\grav\robot+\grav\load$, the closed-loop dynamics of the controlled system becomes, 
\begin{align}   
    \mass \ddot\state+\cori\dot\state - \dsu & =  \textcolor{red}{\uh'}
\end{align} 
\nadiainline{Plugging in Eq. \eqref{eq:u_ds}, Eq. \eqref{eq: Apest} and \eqref{eq: Arest} then it becomes, 
\begin{align}
\label{eq: controlled-closed-dynamics}
    \resizebox{\hsize}{!}{$\mass \left[\begin{array}{l}\ddot\xpos \\ \dot\rotv\end{array}\right]+\cori\left[\begin{array}{l}\dot\xpos \\ \rotv
        \end{array}\right] + \left[\begin{array}{l}
        \impD_{p}(c_{p}(t))\left(\dot{\xpos}-\dsf_\pos(\xpos; \est{\pfx_\pos}) \right) \\
        \impD_{o}(c_{o}(t))\left(\rotv-\dsf_\rot(\quat;\est{\pfx_\rot})\right) 
        \end{array}\right]= \textcolor{red}{\uh'}$}
\end{align} 

\noindent \textit{\textbf{Note:} For simplicity of explanation the following analysis considers only the Cartesian control term (top block of $\dsu$), hence we analyze the following part of Eq.~\eqref{eq: controlled-closed-dynamics}, 
\begin{align}
\label{eq:closed_pos_final}
    \mass \ddot\xpos+\cori\dot\xpos + \impD_{p}(c_{p}(t))\dot{\xpos} - \impD_{p}(c_{p}(t))\dsf_\pos(\xpos; \est{\pfx_\pos})  =  \textcolor{red}{\uh'}
\end{align}
yet, is applicable to the rotational control term as well.}

To analyze the behaviour of our closed-loop system in Eq. \eqref{eq:closed_pos_final}, we derive the apparent damping and stiffness during interaction as in the classical impedance control formulation from Hogan \cite{Hogan1984} following \cite{khoramshahi2019dynamical,lags2022} and stated next. 
\subsubsection{On Damping} First, for the apparent damping matrix $\mathbf{D}_p^h\in\mathbb{R}^{3\times 3}$; i.e., the damping experienced by the user during interaction/contact, we can compute the partial derivative of \textcolor{red}{$\uh'$} from Eq.~\eqref{eq:closed_pos_final} wrt. $\dot\xpos$ as, 
\begin{equation}
    \mathbf{D}_p^h(c_p(t)) = \frac{\partial \uh'}{\partial\dot\xpos} = \cori + \frac{\partial \left(\impD_{p}(c_{p}(t))\dot\xpos\right)}{\partial\dot\xpos},
\end{equation}
assuming the Coriolis terms of the robot are compensated by the low-level torque controller and from the object neglected, the apparent damping of our controlled system becomes
\begin{equation}
\label{eq:closed_damping}
\begin{aligned}
    \mathbf{D}_p^h(c_p(t)) & = \frac{\partial \left(c_{p}(t)\mathbf{\Lambda}_p\dot\xpos\right)}{\partial\dot\xpos} \\
    & = \frac{\partial c_{p}(t)}{\partial\dot\xpos}(\mathbf{\Lambda}_p \dot\xpos)^\top +  c_{p}(t)\mathbf{\Lambda}_p.
    \end{aligned}
\end{equation}
The first term in Eq.~\eqref{eq:closed_damping} can be computed as follows, 
\begin{equation}
\label{eq:partial_conf}
\frac{\partial (c_{p}(t))}{\partial\dot\xpos} = \int_{-T}^0 - \frac{\partial (e_{p}(s))}{\partial\dot\xpos}\dd{s} =\int_{-T}^0 \frac{\est{\dot\xpos}(s)-\dot\xpos(s)}{\twonorm{\est{\dot\xpos}(s)-\dot\xpos(s)}}\dd{s}
\end{equation}
One can then extract the element-wise terms from Eq.~\eqref{eq:partial_conf} $\frac{\partial (c_{p}(t))}{\partial\dot\xpos} = [\frac{\partial (c_{p}(t))}{\partial\dot\xpos_1},\frac{\partial (c_{p}(t))}{\partial\dot\xpos_2},\frac{\partial (c_{p}(t))}{\partial\dot\xpos_3}]^T$, and compute  the outer product $* = \frac{\partial c_{p}(t)}{\partial\dot\xpos}(\mathbf{\Lambda}_p \dot\xpos)^\top$ producing the rank-1 matrix,  
\begin{equation}
\label{eq:closed_damping_Jac}
    * =  \begin{bmatrix}
    d_{p}^1\frac{\partial (c_{p}(t))}{\partial\dot\xpos_1}\dot\xpos_1 & d_{p}^1\frac{\partial (c_{p}(t))}{\partial\dot\xpos_2}\dot\xpos_1 & d_{p}^1\frac{\partial (c_{p}(t))}{\partial\dot\xpos_3}\dot\xpos_1 \\
    d_{p}^2\frac{\partial (c_{p}(t))}{\partial\dot\xpos_1}\dot\xpos_2 & d_{p}^2\frac{\partial (c_{p}(t))}{\partial\dot\xpos_2}\dot\xpos_2 &  d_{p}^2\frac{\partial (c_{p}(t))}{\partial\dot\xpos_3}\dot\xpos_2\\
    d_{p}^3\frac{\partial (c_{p}(t))}{\partial\dot\xpos_1}\dot\xpos_3 & d_{p}^3\frac{\partial (c_{p}(t))}{\partial\dot\xpos_2}\dot\xpos_3 & d_{p}^3\frac{\partial (c_{p}(t))}{\partial\dot\xpos_3}\dot\xpos_3 \\
    \end{bmatrix}.
\end{equation}
Hence, the resulting damping matrix from Eq. \eqref{eq:closed_damping} is composed of diagonal terms $d_{p}^i\left(c_{p}(t)+\frac{\partial (c_{p}(t))}{\partial\dot\xpos_i}\dot\xpos_i\right)$ reflecting both the direct scaling by 
$c_p(t)$ and the contribution from differentiating 
it. The off-diagonal elements are derivatives of $c_p(t)$ with respect to velocity directions other than the one matched by the row, reflecting the contribution of the change in confidence wrt. to the velocities of the robot in each direction. As shown, the apparent damping of our system varies wrt. $c_{p}(t)$ which in turn varies linearly wrt. the Cartesian velocity tracking error $\esterr_{p}(t) = \twonorm{\est{\dot\xpos}-\dot\xpos}$ with $c_{p}(t) = \int_{-T}^0 (\decay_{p}- e_{p}(s)) \dd{s}$ and the velocity of the robot. 

\subsubsection{On Stiffness} Although our proposed variable impedance controller defined in Eq.~\eqref{eq:u_ds} has no explicit stiffness term, like in classical impedance control, one can derive the apparent stiffness $\mathbf{K}_p^h\in\mathbb{R}^{3\times 3}$, by computing the partial derivative of $\uh'$ from Eq.~\eqref{eq:closed_pos_final} wrt. $\xpos$ as follows (as in \cite{khoramshahi2019dynamical,lags2022}), 
\begin{align}
\label{eq:apparent_stiffness}
    \mathbf{K}_p^h(c_{p}(t)) & = \frac{\partial \uh'}{\partial \xpos}= -\impD_{p}(c_{p}(t))\frac{\partial \dsf_\pos(\xpos; \est{\pfx_\pos})}{\partial \xpos} \nonumber\\
    & = -\impD_{p}(c_{p}(t)) \est\dsAp  \\ \nonumber
    & = - c_{p}(t)\mathbf{\Lambda}_p\est\dsAp
\end{align}
as shown, the classical notion of stiffness in our controller depends not only on the variable damping matrix $\impD_{p}(\cdot)$ but also on the estimated linear DS matrix $\est\dsAp$. Hence, our controller exhibits different stiffness in different directions, which can be computed wrt. chosen coordinates as in \cite{khoramshahi2019dynamical, lags2022}.

\subsubsection{Effect of varying $c_p(t)$} Let us now analyze the effect of the varying confidence function $c_p(t)$ on both the apparent damping $\mathbf{D}_p^h(c_p(t))$ and stiffness $\mathbf{K}_p^h(c_p(t))$ matrices. 

\textit{Case 1:} When $c_p(t) = 0$ this means that the velocity tracking error $e_p(t) \uparrow$ is very high and particle filter is not confident. Hence, both matrices are nullified; i.e., 
\begin{align}
\mathbf{K}_p^h(0)  =  \mathbf{0},~\mathbf{D}_p^h(0) =  \mathbf{0}
\end{align}
which results in pure gravity compensation.

\textit{Case 2:} When $0 < c_p(t) < 1$ this means that the velocity tracking is $0 < e_p(t) < d_p $ and particle filter has some confidence about it's estimate (either decreasing or increasing). Here, the apparent damping and stiffness matrices become, 
\begin{align}
    \mathbf{K}_p^h(c_p(t)) & = -c_{p}(t) \mathbf{\Lambda}_p \est\dsAp \\
    \mathbf{D}_p^h(c_p(t)) & = c_{p}(t) \mathbf{\Lambda}_p + \frac{\partial c_{p}(t)}{\partial\dot\xpos}(\mathbf{\Lambda}_p \dot\xpos)^\top
\end{align}
In this case, we can see that stiffness remains a symmetric positive definite matrix $\mathbf{K}_p^h(c_p(t))\succ 0$. For the damping term since $* = \frac{\partial c_{p}(t)}{\partial\dot\xpos}(\mathbf{\Lambda}_p \dot\xpos)^\top$ is a rank-1 matrix, its eigenvalues are $\lambda(*) = \{0,(\mathbf{\Lambda}_p \dot\xpos)^\top\frac{\partial c_{p}(t)}{\partial\dot\xpos}\}$. Since, $c_{p}(t)\mathbf{\Lambda}_p\succ 0$ then $\mathbf{D}_p^h(c_p(t))\succ 0$ iff $\lambda_{max}(c_{p}(t)\mathbf{\Lambda}_p)>(\mathbf{\Lambda}_p \dot\xpos)^\top\frac{\partial c_{p}(t)}{\partial\dot\xpos}$. Geometrically, this means that the angle between Eq. \eqref{eq:partial_conf} and the current motion of the robot must be $\leq \pi$ which always holds as we constantly re-estimate $\est{\dot\xpos}$ to follow human guidance.

\textit{Case 3:} When $c_p(t) = 1$ this means that the velocity tracking error $e_p(t) = 0 $ and the particle filter is very confident. Here, the apparent damping and stiffness matrices become, 
\begin{align}
    \mathbf{K}_p^h(1) = -\mathbf{\Lambda}_p \est\dsAp, ~
    \mathbf{D}_p^h(1) = \mathbf{\Lambda}_p.
\end{align}
In this case, the apparent damping term is equivalent to the damping gain of the variable impedance controller, equivalent to the classic notion of damping from Hogan \cite{Hogan1984}. On the other hand, the stiffness is a constant matrix depending on the product of the constant damping gains and the DS eigenvalues. Both of which are clearly symmetric positive definite matrices.

\subsubsection{Energy Analysis} To study the stability of the physical interaction between the human and the robot, we use the concept of passivity which focuses on analyzing the energy dissipation and power exchange of the closed-loop controlled system following the analyses prior works using DS-based approaches for pHRC \cite{passiveDS,khoramshahi2019dynamical,khoramshahi2020dynamical}. In Appendix \ref{sec:energy_analysis} we provide an extended energy analysis of this closed-loop controlled system under the assumption of perfect control wrench tracking; i.e., the low-level control torques perfectly track the desired end-effector wrench. Next we summarize the two important results. 

First, in the absence of human guidance, when $\textcolor{red}{\uh'} = \zerovec$ our closed-loop controlled system is stable when $\dsu$ is feasible. Second, when $\textcolor{red}{\uh'} \neq  \zerovec$, the system is always passive wrt. the input/output pair $(\uh',\dot\state)$ for $c_p(t) = 0$, $c_p(t) = 1$ and in the case when the confidence is changing, i.e., $0 < c_p(t) < 1$  with $\dot{c}_p(t)>0$ or $\dot{c}_p(t)<0$, then it remains passive only when the following condition holds: 
\begin{equation}
\label{eq:confidence_bound_stability_main}
    - \underbrace{c_{p}(t)}_{>0} \underbrace{\dot\state^T\mathbf{\Lambda}_p\dot\state}_{E_d > 0} \leq \frac{\dot{c}_p(t)}{2}\underbrace{(\state - \est\state^*)^T\mathbf{\Lambda}_p \est\dsAp(\state - \est\state^*)}_{E_p <0},
\end{equation}
with $E_d = \dot\state^T\mathbf{\Lambda}_p\dot\state$ representing the dissipative energy term of the system and $E_p = (\state - \est\state^*)^T\mathbf{\Lambda}_p \est\dsAp(\state - \est\state^*)$ the potential energy term driven by the DS parameter estimates. As shown, when the confidence is decreasing, $\dot{c}_p(t)<0$, the inequality holds, rendering the robot passive to the human interaction forces. However, when the confidence is increasing, $\dot{c}_p > 0$, the robot stiffens up and can loose passivity in a short time interval. This corresponds to prior works showing the unavoidable temporary injection of energy in the system when stiffness increases in classical variable impedance control laws \cite{kronander2016stability}. As discussed in prior human guidance works such behavior is desirable as the robot becomes a pro-active leader when confident \cite{khoramshahi2020dynamical} and it is natural for it to stiffen up for a short time interval. Interestingly, in our formulation notice that such temporary loss of passivity is driven by the value defined for the ascent rate $d_p$ as $\dot{c}_{p}(t)$ is, 
\begin{equation}
\label{eq:confidence_dt}
\dot{c}_{p}(t) = \frac{d}{dt}\int_{-T}^0 (\decay_{p}- e_{p}(s)) \dd{s} = \decay_{p} - e_{p}(t).
\end{equation}
Hence, the smaller the ascent rate for the confidence the shorter the period of time that passivity is lost and vice-versa. Further, we can also express the upper bound of $d_p$ that preserves passivity by re-writing Eq. \eqref{eq:confidence_bound_stability_main} as $-c_{p}(t)E_d \leq \frac{\dot{c}_p(t)}{2}E_p$ and plugging in Eq. \eqref{eq:confidence_dt} which yields, 
\begin{equation}
\label{eq:dp_bound_passivity}
\begin{aligned}
 -2c_{p}(t)\frac{E_d}{E_p} & \leq \dot{c}_p(t) = \decay_{p} - e_{p}(t).
\end{aligned}
\end{equation}
As shown, one can tune $d_p$ based on user preference and the ratio of dissipative/potential energy and tracking error. In Appendix \ref{sec:energy_analysis} we show that the injection of positive energy in this confidence transition time interval is bounded and empirical validation from real-world experiments that when $d_p=0.41$ the robot stiffens for a short time interval of 0.75s. This analysis holds for the rotational motion as well.}

\subsection{Capability-Aware Control: Robot Motion Feasibility}
\label{sec:robot-control}
Since the robot only does gravity compensation when the confidence is low, there are cases where robot does not restrict itself from joint limits, even if in \ref{sec:pf-trimming} we ensured the motion \& goal feasibility when the robot is confident.
Therefore, a real-time scheme is needed to ensure the robot does not exceed joint limits with the direct output from the Cartesian impedance controller. To achieve this, we borrowed ideas from a classic approach, FIRAS force \cite{khatib1986real}, that had been applied to controlling the Kuka iiwa14 manipulator \cite{munoz2018operational}. To respect the joint position limit constraints, a repulsive potential will be produced to resist the violations. Each joint has a lower limit $\underline{\theta}$ and an upper limit $\overline{\theta}$. We calculate the distance between the current joint configuration $\theta$ and the joint limits
\begin{align}
    \Delta\underline{\theta} = \theta - \underline{\theta}, \\
    \Delta\overline{\theta} = \theta - \overline{\theta}.
\end{align}

The repulsive torque for each joint $\ujr$ will only be activated when inside the lower and upper safety margins defined by $\underline{\delta}$ and $\overline{\delta}$, respectively. 
When activated, the repulsive torques are defined as 
\begin{align}
    \underline{\ujr}&= \begin{cases}\eta_9\left(\frac{1}{\Delta\underline{\theta}}-\frac{1}{\underline{\delta}}\right) \frac{1}{{\Delta\underline{\theta}}^2}, & \:\,\,\,\text { if } \Delta\underline{\theta} \leq \underline{\delta} \\ 0, & \:\,\,\,\text { if } \Delta\underline{\theta}>\underline{\delta}\end{cases}, \\
    \overline{\ujr}&= \begin{cases}-\eta_9\left(\frac{1}{\Delta\overline{\theta}}-\frac{1}{\overline{\delta}}\right) \frac{1}{{\Delta\overline{\theta}}^2}, & \text { if } \Delta\overline{\theta} \leq \overline{\delta} \\ 0, & \text { if } \Delta\overline{\theta}>\overline{\delta}\end{cases},
\end{align}
where $\eta_9$ is a gain that defines the rate of change for the repulsive torque $\ujr$ when approaching the joint limits. Combining these for each joint, we obtain $\boldsymbol{\ujr}$.

For intuitive interaction and perceived safety, the robot should try to be kept in a familiar joint configuration $\jntv_N$ during motions for the human. With the 7 DoF robot, such a requirement can be achieved with a secondary task torque, defined as
\begin{align}
    \ujn = -(\jntv - \jntv_N) - \dot{\jntv},
\end{align}
which can then be projected into the nullspace with the projection matrix
\begin{align}
    \mathbf{N} = (\eye - \jacob\T \jacob\T^\dag).
\end{align}

The total torque commands $\uj$ with joint position limits and null space desired configuration is
\begin{align}
    \uj = \boldsymbol{\ujr} + \mathbf{N}\ujn + \jacob\T(\jntv) \textcolor{orange}{\ur\des}.
\end{align}
and the corresponding task space output is $\textcolor{green}{\ur} = \jacob\T^\dag(\jntv) \uj$.

\section{Experiments}
\label{sec:experiments}
\subsection{Experimental Setup}
Since we model intent as a DS towards a goal pose, we created a simple game of pose reaching: 
The human and the robot hold a heavy load (4.5kg) together.
They start in the middle of the workspace and are asked to move the object towards a 3D goal pose in longitudinal ($x$), lateral ($y$) positions, and a specific roll ($\phi$) angle.
When the team reaches the goal pose, the task ends and the team will return to the middle of the workspace for the next goal pose to appear. 
During the entire process, the robot does not know the ground truth goal pose. Fig. \ref{fig: experiment} shows the experimental setup.
The goal pose is displayed on the TV so the human partner can easily see it.

\begin{figure}
  \begin{subfigure}{0.49\columnwidth}
  \includegraphics[width=\textwidth]{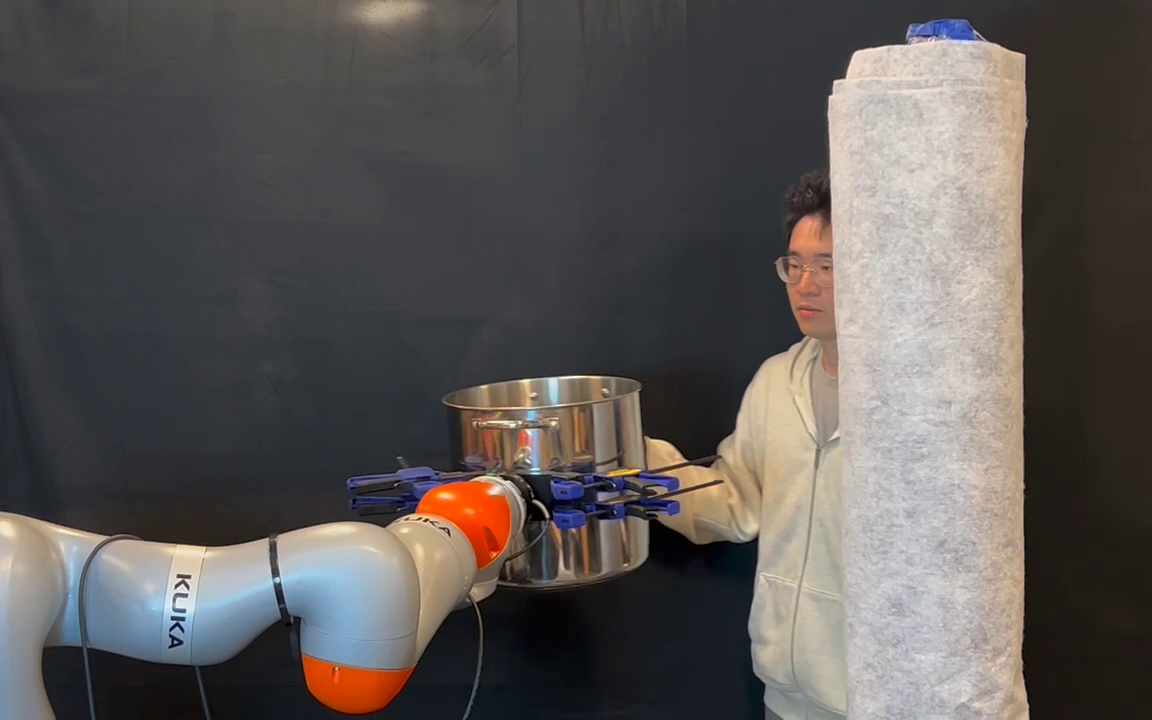}
  \caption{t = 0.0 s}
  \end{subfigure}
  \hfill
  \begin{subfigure}{0.49\columnwidth}
  \includegraphics[width=\textwidth]{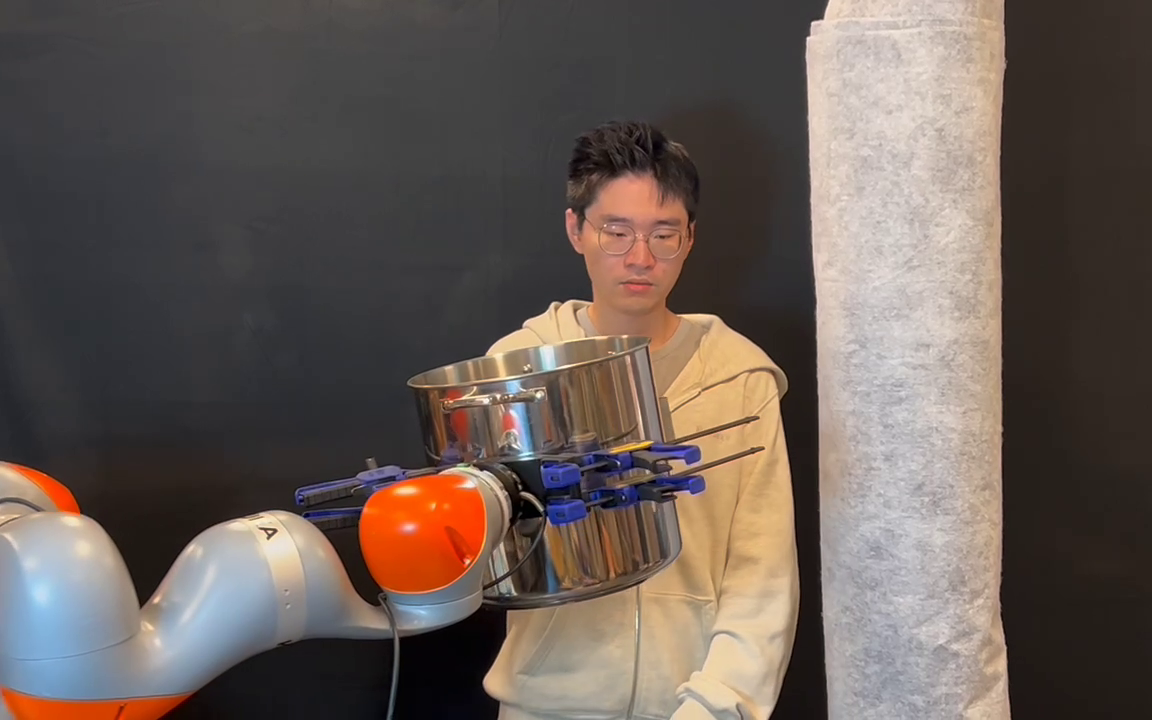}
  \caption{t = 3.0 s}
  \end{subfigure} 
  \begin{subfigure}{0.49\columnwidth} 
  \includegraphics[width=\textwidth]{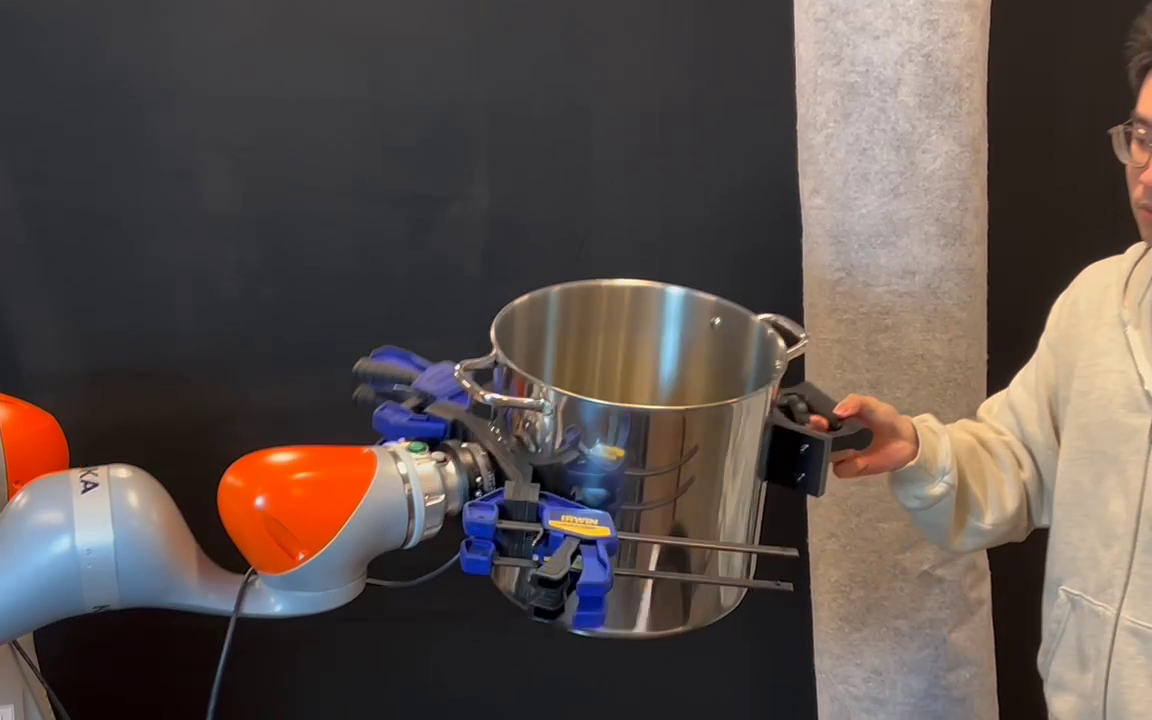} 
  \caption{t = 5.0 s}
  \end{subfigure}  
  \hfill 
  \begin{subfigure}{0.49\columnwidth} 
  \includegraphics[width=\textwidth]{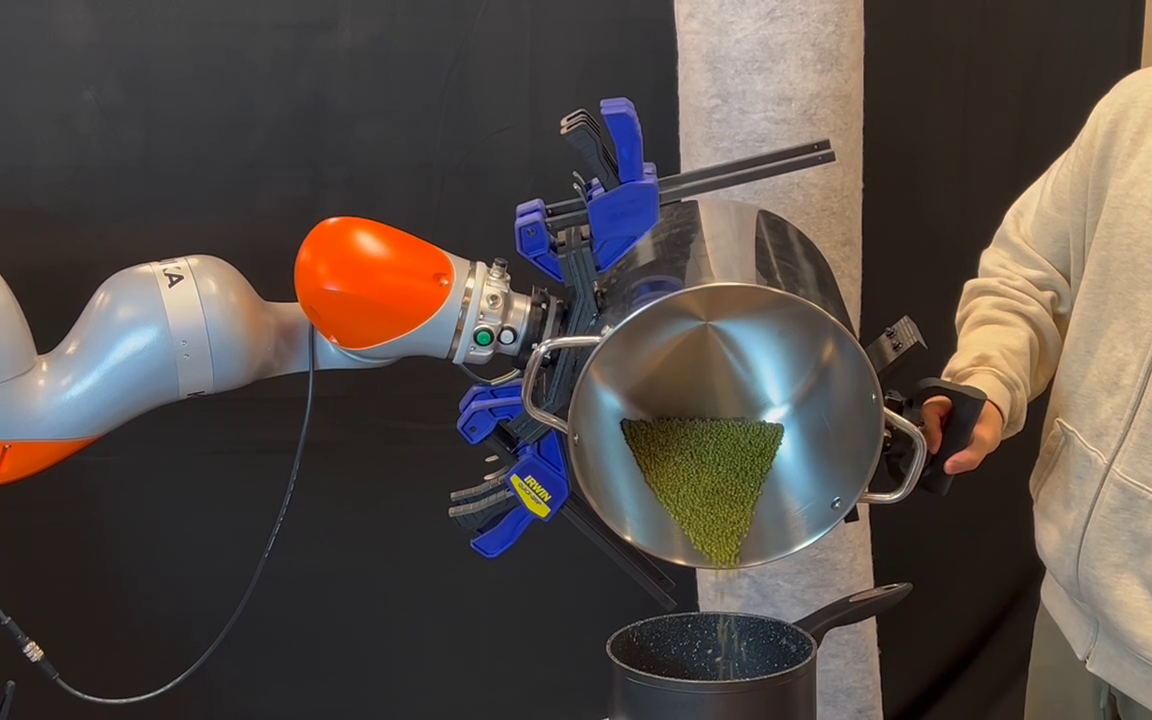} 
  \caption{t = 8.0 s}
  \end{subfigure}
  \caption{This demonstration showcase the proposed method in 6D. The operator has to move the content in the steel pot into another pot, with an obstacle in the way. In (a) and (b), the estimated $x,y$ and yaw goal changes as a result of external input in these dimensions. In (c) and (d), the estimated $z$ and roll goal changes as the operator pushes down and rotate the pot.}
\label{fig: 6d_demo}
\end{figure}

We implemented the proposed method in 6 DoF with two particle filters. A video of the proposed method working in full 6D can be found on the website and the time-lapse is shown in Fig. \ref{fig: 6d_demo}. For all experiments, however, we restrict the particles to previous mentioned task dimensions, $x,y,\phi$. All other dimensions are controlled to predefined goals: $z$ is controlled to a height of 0.3m with a high impedance gain; pitch is controlled to 0 deg; and yaw is controlled to be always pointing outwards from the robot base. For the rotational filter, we estimate the one dimensional dynamics matrix for $\phi$ and apply it to all three rotational dimensions.

To eliminate differences in the controller, we use the proposed variable impedance controller for our method and the same controller with a fixed high impedance gain for all baseline methods. As discussed, this controller comes with features including joint limit constraints and null space controller to nominal joint configuration as the secondary task. The hardware platform we use is a Kuka-iiwa-14 robot.

We carefully perform system identification on the object and perform proper gravity compensation both for the controller and the force sensor when it is on the robot side.
\subsection{Baselines}
We compare our method with three baselines:
\begin{enumerate}
    \item A standard \emph{admittance controller} with force-torque (F/T) sensor on the robot side.
    \item \emph{Task adaptation} \cite{khoramshahi2020dynamical}  with F/T sensor on robot side.
    \item \emph{Task adaptation} \cite{khoramshahi2020dynamical} with F/T sensor on human side.
\end{enumerate}

We provide the motivation for picking these baselines:
\emph{Admittance controller} has been regarded as the gold standard when robots needs to be interacting with humans by directly translating force into motion by simulating a mass damping system for the end effector. However, it often needs to be tuned to work well, and it's performance is not clear in co-manipulation scenarios. For our experiment, we tuned the admittance controller extensively with the load attached so that it feels intuitive. Since our method estimates intent as a motion and a goal, and is aware of the constraints, it should perform better than admittance, which only estimates motion.

\emph{Task Adaptation} adapts in between a few known goals and autonomously goes to the most likely goal based on current velocity. Furthermore, it also uses an admittance component and decides in between tracking human input or one of the predefined goals. Therefore, the human only has to provide the force required to start the motion towards the goal. In our game, we set the two \textit{known} attractor poses including the home pose in the middle of the workspace and the \textbf{ground truth} goal pose. 
This method should be a upper bound on what our method can achieve. To investigate if there is a significant difference when the F/T sensor is on the load side vs. the human side, we also report result of \emph{task adaptation} with the unrealistic setting of having the force sensor on the human side.

The following details are provided for the implementation of the baselines.
\textit{Admittance Controller} generate desired motion of an end-effector given an external wrench to a robotic arm end-effector. It is most commonly used on robots with velocity/position control and the end effector must be equipped with a force/torque sensor.
For our torque-controlled robot, we `simulate' such a robot on hardware by using \textit{admittance control} on top of a stiff impedance controller, where the impedance controller gains are set to be equal to the highest setting of the variable impedance controller in our method.

The \textit{task adaptation} method proposed in \cite{khoramshahi2020dynamical} does not have an open-source implementation. Therefore, we used the implementation of \cite{khoramshahi2019dynamical} and implemented the human admittance controller as proposed. On a high level, this method generates the desired velocity and uses a velocity control interface for control. The desired velocity $\dot\xpos\des = (1-h)\dot\xpos^{\text{t}} + \dot\xpos^{\text{a}} $ is a weighted sum of the task and admittance velocity. $\dot\xpos^{\text{t}}$ is generated by an adaptation mechanism that computes the similarity of the current velocity to the velocity of a few predefined tasks, where each task is a DS. On the other hand, $\dot\xpos^{\text{a}}$ is generated by an admittance controller using the human wrench input and a power pass filter. The filter also informs the controller about the human involvement $h$ and decides the influence of $\dot\xpos^{\text{t}}$. This work is proposed for Cartesian dimensions and we extended it to 6 DoF.
Baseline hyperparameters for both methods are reported in Table \ref{tb: baseline_gain} in Appendix \ref{app: implement}.

\begin{figure}[!t]
    \centering
    \includegraphics[width=0.4\textwidth]{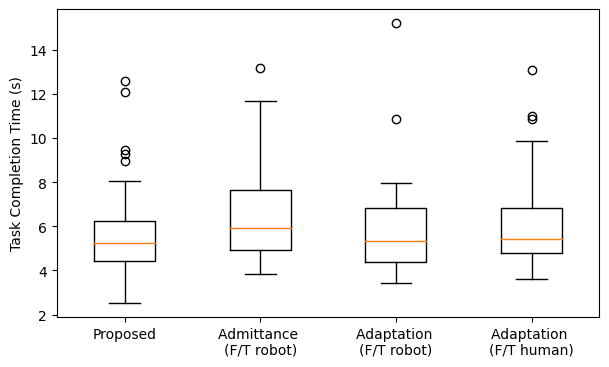}
    \caption{Box plots of task completion time (seconds) over all trials, for the proposed method and baselines. The metric is summarized by the min and max values (black lines), first and third quartiles (box), median (orange line), and outliers (circle).}
    \label{fig: completion_time}
    \vspace{-5pt}
\end{figure}

\subsection{Results}

In the experiment of co-carrying the object towards a goal location. We use \textbf{Task Completion Time}, and \textbf{Human Effort} in terms of impulse, force and torque, to assess the efficiency of the collaboration. 
The task completion time is defined as the time it takes to reach within 0.13m/35deg of the goal, with combined velocity (linear, angular) velocity norm less than 0.1. The data recorded from the force/torque sensor on the human handle is used to extract the linear and angular impulse (i.e., integral the force and torque over time), and the average force and torque applied by the human during each trial.

\begin{figure}[!t]
    \centering
    \includegraphics[trim={.5cm 0 0 0},clip, width=0.495\textwidth]{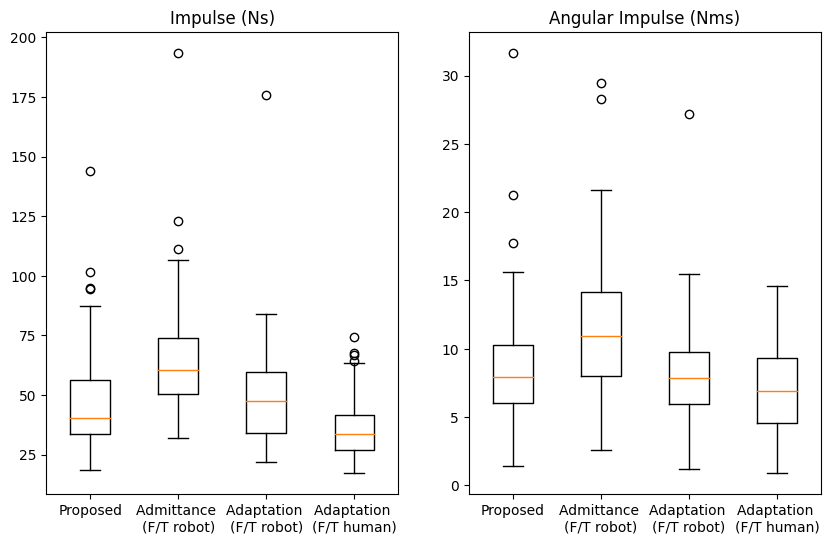}
    \caption{Comparison of \textbf{(left)} linear impulse (Ns) and \textbf{(right)} angular impulse (Nms) over all trials, for the proposed method and baselines, by integration of the readings from F/T sensor on the human side. The box plots show the min and max values (black lines), first and third quartiles (box), median (orange line), and outliers (circle).}
    \label{fig: completion_impulse}
    \vspace{-10pt}
\end{figure}

Three subjects participated in the experiments, and data for 20 trials (with 20 goal poses) is recorded for each method. 
The mean performance metrics over all 20 trials for each subject are reported in Tab. \ref{tab:subjects_results}.
To analyze the overall performance of the methods, the metrics are evaluated on the trials done by all subjects combined. 
Figs. \ref{fig: completion_time} shows the box plots for task completion time. Based on the results, the proposed method requires 83\%, 97\% and 92\% of The time required to complete the task compared to \emph{admittance}, \emph{task adaptation} with F/T sensor on robot side and human side, respectively. The Impulse and Angular Impulse for task completion are shown in Fig. \ref{fig: completion_impulse}. The proposed method requires 76\%, 99\% and 138\% of the impulse, and 77\%, 104\% and 127\% of the angular impulses compared to the baselines. As shown in Fig. \ref{fig: completion_force} the average human effort in the proposed method is 80\%, 88\% and 157\% in terms of force, and 71\%, 96\% and 126\% in terms of torque.

\begin{table}[!ht]
\centering
\caption{Results for all subjects: Mean task completion time, linear impulse, angular impulse, average force (N) and average torque (Nm) through all trials}
\vspace{-5pt}
\footnotesize
\renewcommand{\arraystretch}{1.05}
\begin{tabular}{cc|c|c|c|c|}
\cline{3-6}
& & \multicolumn{4}{c|}{Method} \\ \hline
\multicolumn{1}{|c|}{\multirow{2}{*}{Metric}} &
\multicolumn{1}{c|}{\multirow{2}{*}{{\scriptsize Subj.}}} &
\multicolumn{1}{c|}{\multirow{2}{*}{{Proposed}}} & {\scriptsize Admittance} & {\scriptsize Adaptation} & {\scriptsize Adaptation} \\
\multicolumn{1}{|c|}{} & \multicolumn{1}{c|}{} & \multicolumn{1}{c|}{{}} & \multicolumn{1}{c|}{{\scriptsize F/T robot}} & \multicolumn{1}{c|}{{\scriptsize F/T robot}} & \multicolumn{1}{c|}{{\scriptsize F/T human}} \\ \cline{1-6}
\multicolumn{1}{|c|}{\multirow{3}{*}{Time}} & 
\multicolumn{1}{c|}{1} & \textbf{5.6} & 6.7 & 5.7 & 5.8 \\ \cline{2-6} 
\multicolumn{1}{|c|}{} &
\multicolumn{1}{c|}{2} & 5.2 & 6.2 & \textbf{4.8} & 5.5 \\ \cline{2-6}
\multicolumn{1}{|c|}{(s)} &
\multicolumn{1}{c|}{3} & \textbf{6.0} & 7.2 & 6.7 & 7.0 \\  \cline{1-6}
\multicolumn{1}{|c|}{\multirow{3}{*}{{\scriptsize Lin. Impulse}}} & 
\multicolumn{1}{c|}{1} & 38.9 & 58.0 & 40.9 & \textbf{26.4} \\ \cline{2-6} 
\multicolumn{1}{|c|}{} &
\multicolumn{1}{c|}{2} & 44.3 & 62.6 & 42.9 & \textbf{39.6} \\ \cline{2-6}
\multicolumn{1}{|c|}{(Ns)} &
\multicolumn{1}{c|}{3} & 65.7 & 76.3 & 66.8 & \textbf{42.3} \\ \cline{1-6}
\multicolumn{1}{|c|}{\multirow{3}{*}{{\scriptsize Ang. Impulse}}} & 
\multicolumn{1}{c|}{1} & 7.7 & 10.1 & 6.4 & \textbf{3.9} \\ \cline{2-6} 
\multicolumn{1}{|c|}{} &
\multicolumn{1}{c|}{2} & 8.3 & 13.0 & \textbf{7.6} & 7.7 \\ \cline{2-6}
\multicolumn{1}{|c|}{(Nms)} &
\multicolumn{1}{c|}{3} & 10.7 & 11.8 & 11.6 & \textbf{9.5} \\ \cline{1-6}
\multicolumn{1}{|c|}{\multirow{3}{*}{{\scriptsize Avg. Force}}} & 
\multicolumn{1}{c|}{1} & 6.5 & 9.1 & 7.3 & \textbf{5.0} \\ \cline{2-6} 
\multicolumn{1}{|c|}{} &
\multicolumn{1}{c|}{2} & 6.5 & 9.1 & 7.3 & \textbf{5.0} \\ \cline{2-6}
\multicolumn{1}{|c|}{(N)} &
\multicolumn{1}{c|}{3} & 7.4 & 8.4 & 8.5 & \textbf{3.8} \\ \cline{1-6}
\multicolumn{1}{|c|}{\multirow{3}{*}{{\scriptsize Avg. Torque}}} & 
\multicolumn{1}{c|}{1} & 0.40  & 0.58  & 0.38  & \textbf{0.22} \\ \cline{2-6} 
\multicolumn{1}{|c|}{} &
\multicolumn{1}{c|}{2} & \textbf{0.57}  & 0.95  & 0.66  & 0.60 \\ \cline{2-6}
\multicolumn{1}{|c|}{(Nm)} &
\multicolumn{1}{c|}{3} & 0.66  & 0.64  & 0.62  & \textbf{0.46} \\ \cline{1-6}
\end{tabular}
\label{tab:subjects_results}
\end{table}

\begin{figure}[!t]
    \centering
    \includegraphics[trim={.5cm 0 0 0},clip, width=0.495\textwidth]{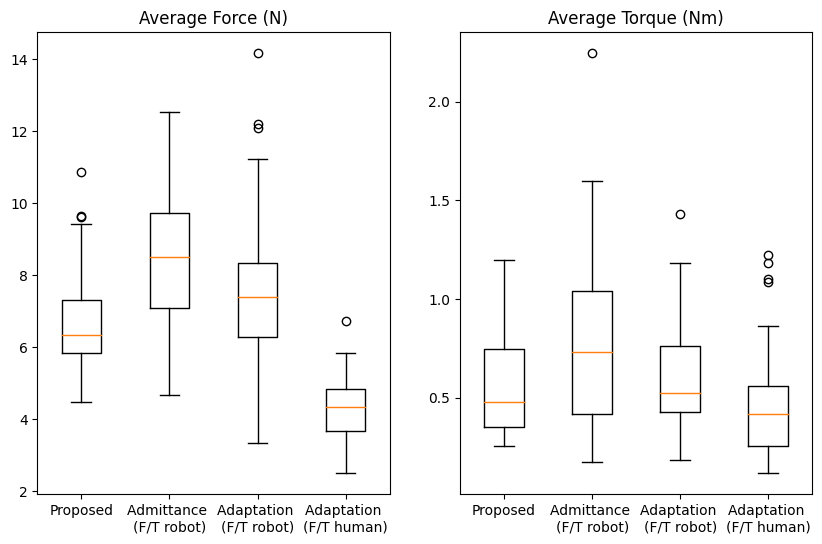}
    \caption{Comparison of \textbf{(left)} average force (N) and \textbf{(right)} average torque (Nm) over all trials, for the proposed method and baselines, as the mean of the readings recorded by the F/T sensor on the human side. The box plots show the min and max values (black lines), first and third quartiles (box), median (orange line), and outliers (circle).}
    \label{fig: completion_force}
    \vspace{-10pt}
\end{figure}

\begin{figure}[ht]
    \centering
    \includegraphics[width=0.50\textwidth]{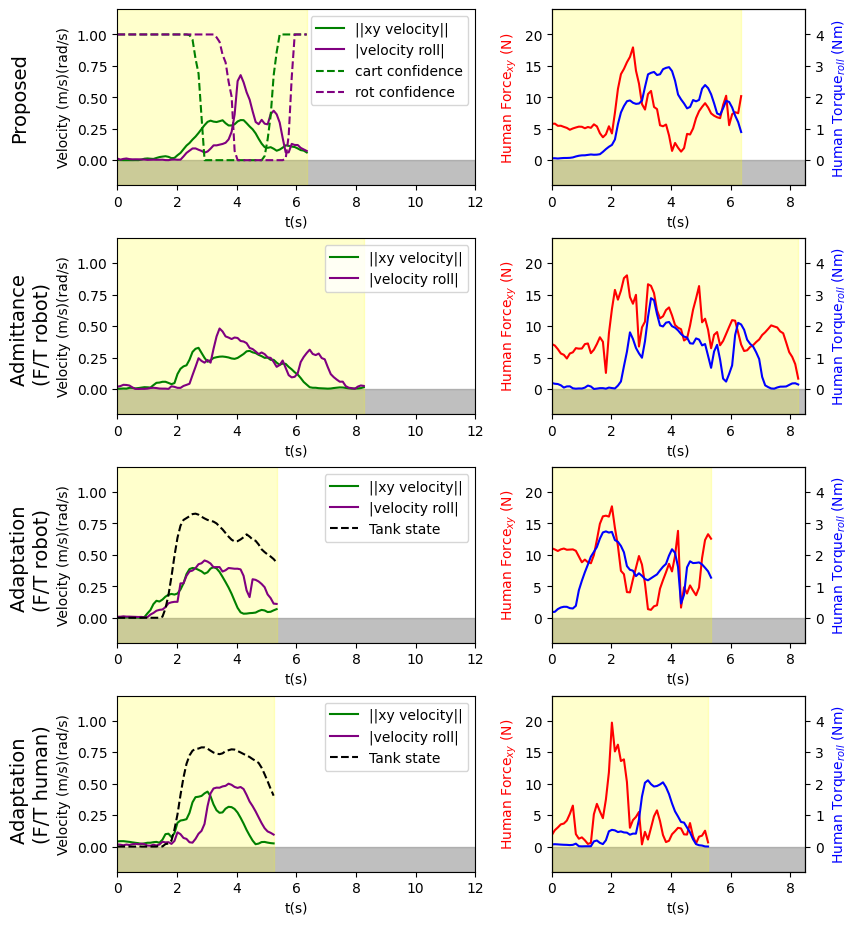}
    \caption{Experiment data of a single trial. Since the task is in 3D space, the left column shows the velocity of the object in $x,y$ directions and for angular velocity on roll angle $\phi$. For the proposed method, we further show confidence, and for Task Adaptation we show the energy tank level. The right column shows the force and torque applied by the human using the force/torque sensor on the human handle. The yellow shade indicates the trial duration (from the time the goal appears on the screen to the time goal is reached)}.%
    \label{fig: trial}
    \vspace{-10pt}
\end{figure}
\begin{figure}[ht]
    \centering
    \includegraphics[trim={1.5cm 1cm 1.5cm 1.5cm},clip, width=0.4\textwidth]{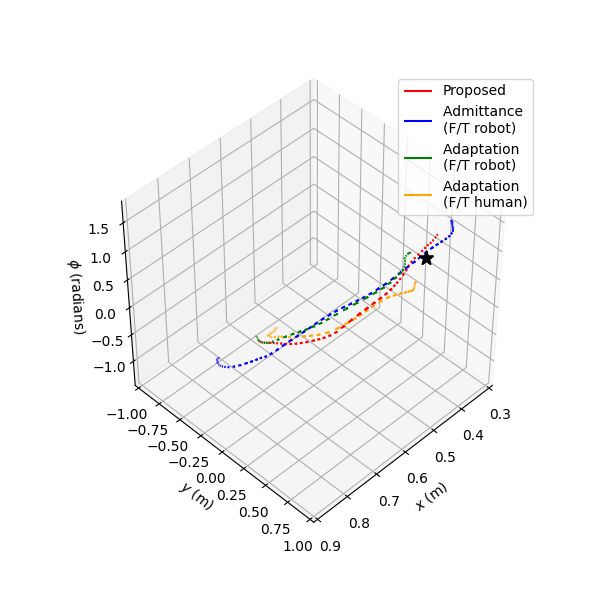}
    \caption{The trajectory of the methods for reaching the same goal location, associated with trial in Fig. \ref{fig: trial}.}
    \label{fig: trial_traj}
    \vspace{-10pt}
\end{figure}

Fig. \ref{fig: trial} shows the typical performance of all four methods and Fig. \ref{fig: trial_traj} shows the trajectory in the three tracked dimensions.
The proposed method generally requires a similar amount of force to activate compared to all baselines, but once activated it requires lower wrench compared to \emph{admittance} to bring a higher velocity, making guiding the robot towards a goal easier. Once the goal is fixed (i.e., around the end of the trial), the confidence increases and the robot automatically resumes the predicted velocity and decelerates towards the goal. This generally leads to shorter task completion times than \emph{admittance}.
\emph{Admittance} generally requires a proportional amount of force or torque to generate a velocity. Notably, one problem in mounting the F/T sensor on the robot side is that since there is a large force and torque from the load, a non-existent force can be measured in the direction towards the human, degrading performance of admittance-based methods.
\emph{Task adaptation} needs the least amount of activation force in general as also seen in Tab. \ref{tab:subjects_results}. However, since it has fixed dynamics matrix towards the goal, in some cases, subjects found it to be too slow and used more effort to force the pot towards the goal pose. The proposed approach addresses this mismatch with the subject expectation by estimating the dynamics matrix online and automatically adapting to each subject's preference.

Overall, all methods alleviate the human effort in lifting the object. 
The proposed method, however, due to the variable impedance controller, outperforms baselines for its higher maneuverability than admittance-based methods. 
All three baselines require the force-torque sensor to be accurate. 
However, under heavy load, even with careful model identification, the force sensor on the robot side experiences non-existent wrenches that need filtering, affecting performance. 
As a result, \emph{admittance} requires constant force input from the human to keep the object still. \emph{Task Adaptation} experiences the same issue but the power filter ignores lower wrenches, making it a better experience. 

\section{Discussion \& Conclusion}
In this paper, we focus on human-robot co-manipulation in a full 6D task space which is a crucial yet underexplored gap in pHRI field, particularly emphasizing the often overlooked angular motion in object-carrying tasks.
We propose a novel approach to modeling intent using Dynamical Systems.
This method captures the continuous nature of intent by leveraging particle filters to estimate the probability distribution of full 6D dynamics matrix and goals intended by the human.
Incorporating the concept of human hand manipulability as a key metric for directional preferences, we take both human intent and task feasibility into consideration.
To ensure a responsive and intuitive collaboration, we additionally introduce a dynamic leader/follower role adaptation to allow the robot to react based on human needs, by real-time adaptation of particle filter noise and impedance control gains.
The demonstrated proof of concept involves a collaborative Human-Robot Collaboration (HRC) task, where the robot seamlessly supports the human counterpart in carrying a 4.5 kg heavy pot. The system adeptly responds to the human's planar motion cues, effectively sharing the load. 
Additionally, the robot provides rotational assistance, thereby alleviating torque on the human wrist. This successful application showcases the potential of our methodology in real-world scenarios, highlighting its practicality and efficiency.
In the conducted comparison, our method outperformed the admittance baseline in a 3D pose-reaching game. Participants subjectively reported a notable reduction in physical strain and a smoother collaboration experience when engaging with our approach. Task adaptation, even with the ground truth goal, performs similarly to our the proposed method.

A key advantage of our method is its closed-loop stability and passivity when no human is interacting with it, detailed in Section \ref{sec:conf-vic}.
However, this does not come at the cost of reduced liveliness since the prediction of the dynamics enables the robot to reach afar with limited interaction time with the human. However, at the moment, to ensure feasibility, the goals are limited to the workspace of the robot, limiting the ranges of the dynamics matrix. 
 Furthermore, the confidence ascent rates need to be tuned so the robot achieves a balance of disturbance rejection and maneuverability. In future work, we hope these can be adjusted dynamically online. Lastly, we hope to extend this work to a mobile manipulator, allowing collaborative transport over long distances with guarantees.

\newpage
\section*{Acknowledgments}
We gratefully acknowledge the support of The Institute for Learning-Enabled Optimization at Scale (TILOS) funded by the National Science Foundation (NSF) under NSF Grant CCR-2112665, IoT4Ag ERC funded through NSF Grant EEC-1941529.

\clearpage
\bibliographystyle{unsrtnat}
\bibliography{references}

\clearpage
\appendix
\subsection{Implementation}
\label{app: implement}
We provide the implementation details of the proposed method.
 Refering to Fig. \ref{fig: ctrl_diagram}, the robot controller (200 Hz) (Capability-Aware Control, DS Variable Impedance Controller) are implemented in C++ using the open-source ROS stack \textit{iiwa\_ros}\footnote{\url{https://github.com/epfl-lasa/iiwa_ros}}.
  The particle filters (20 Hz) are implemented in Python with significant changes on top of \textit{pfilter}\footnote{\url{https://pypi.org/project/pfilter/}}. 
  For capturing the human pose, we use the OptiTrack to obtain the relative rotations between human links and use the captured joint angles and the URDF from \textit{human\_gazebo}\footnote{\url{https://github.com/robotology/human-gazebo}}  to compute the manipulability ellipsoid. 
  Experiments are performed on a desktop with Nvidia GeForce RTX-3070 for Curobo.
  The proposed method hyperparameters are reported in Table \ref{tb: var_gain}. The baselines hyperparameters are reported in Table \ref{tb: baseline_gain}.

\begin{table}[h]
\centering
\begin{tabular}{|l|l|l|}
\hline
                       & Low  & High   \\ \hline
Damping Gain Cartesian $\impD_\pos$ & 1    & 85     \\ \hline
Damping Gain Rotation $\impD_\rot$  & 1    & 13     \\ \hline
Dynamics Cartesian $\dsAp$    & -0.6 & -0.4   \\ \hline
Dynamics Rotation  $\dsAr$     & -0.9 & -0.6   \\ \hline
Relative Weights in Particle Filter $\eta_1 \sim \eta_4$  & 0.5    &      \\ \hline
Ellipsoid Decay Weight $\tune_5$       & 1.5 &    \\ \hline
Noise Dynamics Cartesian $\tune_6$         & 3e-4 & 4e-3 \\ \hline
Noise Dynamics Rotation $\tune_7$         & 2e-4 & 8.5e-3 \\ \hline
Noise Goal Rotation $\tune_8$         & 2e-4 & 8.5e-3 \\ \hline
Joint Limit Resistance $\eta_9$         & 0.1& \\ \hline
Ascent Rate Cartesian $\decay_p$ &  0.41 &\\ \hline
Ascent Rate Rotation $\decay_o$ & 0.49 &\\
\hline
\end{tabular}
\caption{Proposed Method Hyperparameters. Fixed values are written in low column}
\label{tb: var_gain}
\end{table}

\begin{table}[h]
\centering
\begin{tabular}{|l|l|l|l|}
\hline
\cellcolor[HTML]{C0C0C0}Admittance & Value & \cellcolor[HTML]{C0C0C0}Task Adaptation & Value \\ \hline
Mass Cartesian                     & 10.0   & Tank Size                               & 2     \\ \hline
Mass Rotation                      & 2.0   & Trigger                                 & 1     \\ \hline
Damping Cartesian                  & 30.0     & Force Deadzone                          & 15    \\ \hline
Damping Rotation                   & 5.0     & Torque Deadzone                         & 0.8   \\ \hline
Max Velocity                       & 0.8   & Max Velocity                            & 1.3   \\ \hline
Max Acceleration                   & 1.0   & Max Acceleration                        & 2.0   \\ \hline
\end{tabular}
\caption{Baseline Parameters}
\label{tb: baseline_gain}
\end{table}

\subsection{Preliminaries on Quaternions \& Rotational Motion}
\label{app: quat_prelims}
The unit quaternion $\quat\in\mathbb{S}^3\subset\mathbb{R}^4$, where $\mathbb{S}^3$ is the unit hypersphere in $\mathbb{R}^4$, can be defined as, 
\begin{equation}
\label{eq:quat}
\quat = \begin{bmatrix}
s \\
\boldsymbol{u}  \\
\end{bmatrix} = \begin{bmatrix}
\text{scalar}(\quat) \\
\text{vec}(\quat) \\
\end{bmatrix} = \begin{bmatrix}
\cos(\theta/2) \\
\sin(\theta/2)\boldsymbol{n} \\
\end{bmatrix}
\end{equation}
where $\boldsymbol{u}=[u_x,u_y,u_z]^T\in\Rtri$, $s\in\R$ and  $\theta$,$\boldsymbol{n}$ are the angle and normalized axis of rotations in the axis-angle representation, respectively. Note how the functions $\text{scalar}(\cdot)$ and $\text{vec}(\cdot)$ extract the scalar and vector components of a quaternion. The angular velocity $\boldsymbol{\omega}\in\mathbb{R}^3$ required to rotate $\quat_2$ onto $\quat_1$ is computed as follows,
\begin{equation}
\label{eq:omega_geom_quat}
\boldsymbol{\omega} = 2\log(\quat_1 \otimes \mathbf{\bar{q}}_2)/\Delta t
\end{equation}
where for $\Delta\quat = \quat_1 \otimes \mathbf{\bar{q}_2}$, the quaternion logarithm equation  $\log(\cdot):\mathbb{S}^3\rightarrow\mathbb{R}^3$ is computed as follows,
\begin{equation}
\label{eq:log_geom_quat}
\log(\Delta\quat) = \log(\begin{bmatrix}
s \\
\boldsymbol{u}  \\
\end{bmatrix} ) = \begin{cases}
\arccos(s)\frac{\boldsymbol{u}}{||\boldsymbol{u}||} & \text{if} ~ ||\boldsymbol{u}||> 0\\
[0,0,0]^T, & \text{otherwise}
\end{cases}
\end{equation}
which is the geometric logarithm equation that takes into account the singularity at quaternion $\quat=[-1, 0,0,0]^T$ \cite{6907291} and with $\Delta t$ being the time-step. 

Note that the expression $\Delta\quat = \quat_1 \otimes \boldsymbol{\bar{q}_2}$ in \eqref{eq:log_geom_quat} signifies the rotation difference from $\quat_2\rightarrow\quat_1$ and is computed with the following quaternion algebraic operations. 
$
\mathbf{\bar{q}} = \begin{bmatrix}
s\\
-\boldsymbol{u}  \\
\end{bmatrix}
$ is the quaternion conjugate.
$\quat_1 \otimes \quat_2$ represents the product of two quaternions. Given 
$
\quat_1 = \begin{bmatrix}
s_1\\
\boldsymbol{u}_1  \\
\end{bmatrix}, \quat_2 = \begin{bmatrix}
s_2\\
\boldsymbol{u}_2  \\
\end{bmatrix}
$
the partitioned quaternion product is,
\begin{equation}
\label{eq:quat_prod_1}
\quat_1 \otimes \quat_2 = \begin{bmatrix}
s_1s_2 - \boldsymbol{u}_1^T\boldsymbol{u}_2\\
s_1\boldsymbol{u}_2 +  s_2\boldsymbol{u}_1 +  \mathbf{S}(\boldsymbol{u}_1)\boldsymbol{u}_2 \\
\end{bmatrix}.
\end{equation}
with $\mathbf{S}(\boldsymbol{u})\in\mathbb{R}^{3\times 3}$ being a skew-symmetric matrix with the following known properties: (i) $\mathbf{S}(\boldsymbol{u})^T = -\mathbf{S}(\boldsymbol{u})$, (ii) $\mathbf{S}(\boldsymbol{u})\boldsymbol{u}=0$ and (iii) $\mathbf{S}(\boldsymbol{u}_1)\boldsymbol{u}_2=-\mathbf{S}(\boldsymbol{u}_2)\boldsymbol{u}_1$. Thus, the quaternion difference vector is computed as:
\begin{equation}
\label{eq:quat_diff_vector}
\quat_1 \otimes \bar{\quat}_2 = \begin{bmatrix}
s_1s_2 + \boldsymbol{u}_1^T\boldsymbol{u}_2\\
-s_1\boldsymbol{u}_2 +  s_2\boldsymbol{u}_1 -  \mathbf{S}(\boldsymbol{u}_1)\boldsymbol{u}_2 \\
\end{bmatrix}.
\end{equation}
Even though the quaternion product is non-commutative, i.e. $\quat_1 \otimes \bar{\quat}_2 \neq \quat_2 \otimes \bar{\quat}_1$ thanks to the properties of skew-symmetric matrices the following equivalences hold. \\
For the scalar part of the products:
\begin{equation}
\label{eq:scalar_eq}
\begin{aligned}
\text{scalar}(\quat_1 \otimes \bar{\quat}_2) & = \text{scalar}(\quat_2 \otimes \bar{\quat}_1)\\
s_1s_2 + \boldsymbol{u}_1^T\boldsymbol{u}_2 & = s_2s_1 + \boldsymbol{u}_2^T\boldsymbol{u}_1
\end{aligned}
\end{equation}
For the vector part of the products:
\begin{equation}
\label{eq:vector_eq}
\begin{aligned}
\text{vec}(\quat_1 \otimes \bar{\quat}_2) & = -\text{vec}(\quat_2 \otimes \bar{\quat}_1)\\
-s_1\boldsymbol{u}_2 +  s_2\boldsymbol{u}_1 -  \mathbf{S}(\boldsymbol{u}_1)\boldsymbol{u}_2 & = -(-s_2\boldsymbol{u}_1 +  s_1\boldsymbol{u}_2 -  \mathbf{S}(\boldsymbol{u}_2)\boldsymbol{u}_1)\\
 & = -  s_1\boldsymbol{u}_2 + s_2\boldsymbol{u}_1 +  \underbrace{\mathbf{S}(\boldsymbol{u}_2)\boldsymbol{u}_1}_{= -\mathbf{S}(\boldsymbol{u}_1)\boldsymbol{u}_2}
\end{aligned}
\end{equation}
which yields
\begin{equation}
\label{eq:vector_norm_eq}
\begin{aligned}
||\text{vec}(\quat_1 \otimes \bar{\quat}_2)|| & = ||\text{vec}(\quat_2 \otimes \bar{\quat}_1)||
\end{aligned}
\end{equation}
and for the $\log(\cdot)$ error function using \eqref{eq:scalar_eq}, \eqref{eq:vector_eq} and \eqref{eq:vector_norm_eq}:

\begin{equation}
\label{eq:log_eq1}
\begin{aligned}
\log(\quat_1 \otimes \bar{\quat}_2) & = -\log(\quat_2 \otimes \bar{\quat}_1)
\end{aligned}
\end{equation}
\begin{equation}
\label{eq:log_eq2}
\resizebox{\hsize}{!}{$\begin{aligned}
\frac{\arccos(\text{scalar}(\quat_1 \otimes \bar{\quat}_2))}{||\text{vec}(\quat_1 \otimes \boldsymbol{\bar{q}_2})||}\text{vec}(\quat_1 \otimes \boldsymbol{\bar{q}_2}) & = 
- \frac{\arccos(\text{scalar}(\quat_2 \otimes \bar{\quat}_1))}{||\text{vec}(\quat_2 \otimes \bar{\quat}_1)||}
\text{vec}(\quat_2 \otimes \bar{\quat}_1)
\end{aligned}$}
\end{equation}

The evolution of a time-varying unit quaternion with angular velocity $\boldsymbol{\omega}(t)=[\omega_1(t),\omega_2(t),\omega_3(t)]^T$ is defined by the following differential equation,
\begin{equation}
\label{eq:quat_diff}
\dot{\quat} = \frac{1}{2}\boldsymbol{\tilde{\omega}} \otimes \quat \rightarrow \dot{\quat} = \begin{bmatrix}
\dot{s}\\ \boldsymbol{\dot{u}} 
\end{bmatrix} = \begin{bmatrix} -\frac{1}{2} \boldsymbol{u}^T\boldsymbol{\omega}\\
\frac{1}{2} (s\mathbf{I} - \mathbf{S}(\boldsymbol{u}))\boldsymbol{\omega}
\end{bmatrix}
\end{equation}
with $\boldsymbol{\tilde{\omega}} = [0,\boldsymbol{\omega}^T]^T$. This relation between the time derivative of a quaternion and its angular velocity is referred as quaternion propagation \cite{8793786}. Assuming $\boldsymbol{\omega}$ is time independent constant in unit time, one can integrate \eqref{eq:quat_diff} to obtain the next desired quaternion with the quaternion exponential map $\exp(\cdot): \mathbb{R}^3 \rightarrow \mathbb{S}^3$, 
\begin{equation}
\label{eq:quat_int}
\quat(t+\Delta t) = \exp\left( \boldsymbol{\omega}(t)\Delta t/2 \right)\otimes\quat(t).
\end{equation}
The quaternion exponential equation $\exp(\cdot)$ is computed with its modified form to take into account the singularity, analogous to \eqref{eq:log_geom_quat}, as below,
\begin{equation}
\label{eq:exp_geom_quat}
\exp\left( \boldsymbol{\omega}\Delta t/2\right)= \begin{cases}
\begin{bmatrix}\cos(||\boldsymbol{\omega}\Delta t/2||)\\  \frac{\boldsymbol{\omega}}{||\boldsymbol{\omega}||}\sin(||\boldsymbol{\omega}\Delta t/2||) \end{bmatrix}  & \text{if} ~ ||\boldsymbol{\omega}\Delta t/2||> 0\\
[1,0,0,0]^T, & \text{otherwise}
\end{cases}
\end{equation}
Since both equation \eqref{eq:log_geom_quat} and \eqref{eq:quat_int} are limiting the domain of $\boldsymbol{\omega}$ to $0\leq ||\boldsymbol{\omega}||\leq \pi $, they are one-to-one mappings, continuously differentiable and inverse to each other \cite{6907291,UDE1999163}.

\subsection{\revise{Stability Proof of Linear Quaternion DS (Theorem \ref{thm:orient_DS})}}
\label{sec:linear_quatDS}
\noindent \textbf{Proof:} To prove Theorem \ref{thm:orient_DS}, we propose the following Lyapunov candidate function:
\begin{equation}
\label{eq:V_q}
V_\rot(\quat) = V_\rot(\begin{bmatrix}s\\\boldsymbol{u}\end{bmatrix}) = (s^{*} - s)^2 + ||\boldsymbol{u}^{*} - \boldsymbol{u} ||^2
\end{equation}
which is positive definite, radially unbounded and continuously differentiable. To prove global asymptotic stability of \eqref{eq:ds_rot_dot} towards $\quat^*$ the following conditions must hold: (i) $V_\rot(\quat^*) = 0$, (ii) $V_\rot(\quat) > 0 ~ \forall \quat \in \mathbb{S}^3 \setminus \quat = \quat^*$, (iii) $\dot{V_\rot}(\quat^*) = 0$ and (iv) $\dot{V_\rot}(\quat) < 0 ~ \forall \quat \in \mathbb{S}^3 \setminus \quat = \quat^*$. From \eqref{eq:V_q} it is straightforward to see that conditions (i) and (ii) hold. To prove conditions (iii) and (iv) we compute the time derivative of \eqref{eq:V_q} as below, 
\begin{equation}
\label{eq:V_q_dot}
\dot{V_\rot}(\quat) = -2(s^{*} - s)\dot{s} - 2(\boldsymbol{u}^{*} - \boldsymbol{u})^T\boldsymbol{\dot{u}}
\end{equation}

\noindent Using the quaternion propagation relations stated in \eqref{eq:quat_diff} and the properties of skew-symmetric matrices, \eqref{eq:V_q_dot} becomes:
\begin{equation}
\label{eq:V_q_dot_2}
\begin{aligned}
\dot{V_\rot}(\quat) & = -2(s^{*} - s)(-\frac{1}{2}\boldsymbol{u}^T\boldsymbol{\omega}) - 2(\boldsymbol{u}^{*} - \boldsymbol{u})^T\Big( \frac{1}{2}(s_{}\mathbf{I} - \mathbf{S}(\boldsymbol{u})) \boldsymbol{\omega}\Big)   \\
& = \boldsymbol{\omega}^T\Big( \boldsymbol{u}(s^{*} - s) - (s\mathbf{I} - \mathbf{S}(\boldsymbol{u})^T)(\boldsymbol{u}^{*} - \boldsymbol{u})\Big)\\
& = \boldsymbol{\omega}^T\Big(s^{*}\boldsymbol{u} - s\boldsymbol{u} - s\boldsymbol{u}^{*}  + s\boldsymbol{u} + \mathbf{S}(\boldsymbol{u})^T\boldsymbol{u}^{*} - \mathbf{S}(\boldsymbol{u}_{})^T\boldsymbol{u}_{}\Big)\\
& = \boldsymbol{\omega}^T\underbrace{\Big(s^{*}\boldsymbol{u} - s\boldsymbol{u}^{*} - \mathbf{S}(\boldsymbol{u})\boldsymbol{u}^{*}\Big)}_{\text{vec}(\quat \otimes \bar{\quat}^{*})}\\
& = \text{vec}(\quat \otimes \bar{\quat}^{*})^T\boldsymbol{\omega}_{}
\end{aligned}
\end{equation}
Now, substituting \eqref{eq:ds_rot_dot} in \eqref{eq:V_q_dot_2} yields,
\begin{equation}
\label{eq:V_q_dot_3}
\begin{aligned}
\dot{V_\rot}(\quat) & = \text{vec}(\quat \otimes \bar{\quat}^{*})^T\boldsymbol{\omega}_{} \\
& = \text{vec}(\quat \otimes \bar{\quat}^{*})^T\Big( \mathbf{A}_o \underbrace{k_q(\quat,\quat^{*})}_{\text{Via} ~ \eqref{eq:stability_cond_quat}} \underbrace{\log( \quat \otimes \bar{\quat}^{*})}_{\text{Via} ~ \eqref{eq:log_geom_quat}} \Big)\\
& = \text{vec}(\quat \otimes \bar{\quat}^{*})^T\mathbf{A}_o \Big( \frac{||\text{vec}(\quat \otimes \bar{\quat}^{*})||}{\arccos(\text{scalar}(\quat \otimes \bar{\quat}^{*}))}\cdot\\
& ~~~~~~~~~~~~\frac{\arccos(\text{scalar}(\quat \otimes \bar{\quat}^{*}))}{||\text{vec}(\quat \otimes \bar{\quat}^{*})||} \text{vec}(\quat \otimes \bar{\quat}^{*}) \Big)\\
& = \text{vec}(\quat \otimes \bar{\quat}^{*})^T\mathbf{A}_o\text{vec}(\quat \otimes \bar{\quat}^{*})\\
& = -\text{vec}(\quat^* \otimes \bar{\quat})^T\mathbf{A}_o\Big(-\text{vec}(\quat^{*} \otimes \bar{\quat})\Big)\\
& = \text{vec}(\quat^* \otimes \bar{\quat})^T\underbrace{\mathbf{A}_o}_{\prec 0}\text{vec}(\quat^{*} \otimes \bar{\quat}) \leq 0\\
\end{aligned}
\end{equation}
By substituting $\quat = \quat^* $ in \eqref{eq:V_q_dot_3} condition (iii) is ensured, i.e., $\dot{V_\rot}(\quat^*)=0$, and condition (iv) is also ensured, i.e., $\dot{V_\rot}(\quat) < 0 ~ \forall \quat \in \mathbb{S}^3 \setminus \quat = \quat^*$
Hence, the attractor $\quat^*$ is globally asymptotically stable, i.e.,
\begin{equation}
\lim_{t\rightarrow \infty}||\text{vec}(\quat^{*} \otimes \bar{\quat}(t))|| = 0 ~ \implies ~\lim_{t\rightarrow \infty}||\log(\quat^{*} \otimes \bar{\quat}(t))|| = 0 
\end{equation}
if all condition stated in \eqref{eq:stability_cond_quat} are met.
$\hfill \blacksquare$

\subsection{Derivative of Rotational Estimate}
We derive the derivative of \eqref{eq: Arest}. We first define $\ff$ as
\begin{align}
\ff &=  \text{vec}(\quat \otimes \bar{\quat}^*) = -s\quatu^*+s^*\quatu -  \skewsym(\quatu) \quatu^*\\
\end{align}
Then, we first rewrite \eqref{eq: Arest} as
\begin{align}
  \est{\rotv} &=\est\dsAr \ff.\\
\end{align}
We then take the derivative of $\est{\rotv}$ with respect to time as
\begin{align}
  \est{\rota} &=\est\dsAr \dot{\ff},\\
\end{align}
\noindent where $\dot{\ff}$ is given by
\begin{align}
  \dot{\ff} &= \begin{bmatrix}
    -\quatu^* & -s^*\mathbf{I} + \skewsym(\quatu^*)\\
  \end{bmatrix}
  \begin{bmatrix}
    \dot{s}\\
    \dot{\quatu}\\
  \end{bmatrix}\\
  &= \begin{bmatrix}
    -\quatu^* & -s^*\mathbf{I} + \skewsym(\quatu^*)\\
  \end{bmatrix}
  \begin{bmatrix}
    -\frac{1}{2}\quatu^T\rotv\\
    \frac{1}{2}(s\mathbf{I} - \skewsym(\quatu))\rotv\\
  \end{bmatrix}\\
  &= \frac{1}{2}\begin{bmatrix}
    \quatu^*\quatu^T + (-s^*\mathbf{I} + \skewsym(\quatu^*)) (s\mathbf{I} - \skewsym(\quatu))\\ 
  \end{bmatrix} \rotv 
\end{align}

\subsection{\revise{Stability Proof of DS Particle Filter Estimates (Lemma \ref{lemma:stab_pf})}}
\label{app: proofpf}
\noindent\textbf{Proof:} The Cartesian DS proof is trivial and can be found in \cite{billard2022learning}. Here we reproduce for convenience and begin with $\dsf_\pos$.
We choose the following candidate Lyapunov function, 
\begin{align}
V_\pos(\xpos) = \frac{1}{2} (\xpos - \dsgpos)\T (\xpos - \dsgpos).
\end{align} 
Taking derivative w.r.t. time, we have 
\begin{align}
\label{eq:V_p_dot}
    \dot{V_\pos}(\xpos) &= (\xpos - \dsgpos)\T \dot\xpos\\
    &= (\xpos - \dsgpos)\T\underbrace{\dsAp}_{\prec 0} (\xpos - \dsgpos) < 0 ~~ \forall~\xpos \neq \dsgpos
\end{align}
Note that the DS particle filter produces a diagonal matrix that is always negative definite $\dsAp \prec \zerovec$. Hence, $\dot{V_\pos}(\xpos)<0$ expect when $\xpos = \dsgpos \rightarrow \dot{V_\pos}(\dsgpos) = 0$. Since $V_\pos(\xpos) > 0 , V_\pos(\dsgpos) = 0, \dot{V_\pos}(\dsgpos) = 0$ all Lyapunov conditions for global asymptotic stability are satisfied. Hence, the Cartesian DS converges globally to $\dsgpos$; i.e., $\lim_{t\rightarrow\infty}||\xpos - \dsgpos||=0$.

Further, for $\dsf_\rot$ we proved in Theorem \ref{thm:orient_DS} the rotational DS will be globally asymptotically stable if $\dsAr \prec 0$ and $ k_q(\quat,\quat^{*}) = \frac{||\text{vec}(\quat \otimes \bar{\quat}^*)||}{\arccos(\text{scalar}(\quat \otimes \bar{\quat}^{*}))}$. The first condition is satisfied in the particle trimming step: if any element of the estimated diagonal dynamics matrix exceeds the upper or lower bound, the particle receives a weight of 0; The second condition is satisfied by construction. Therefore, the estimated rotational DS defined by $\est{\pfx_\rot}$ is globally asymptotically stable.$\hfill \blacksquare$

\subsection{Energy Analysis of Closed-Loop Controlled System}
\label{sec:energy_analysis}
Following we present the stability and passivity analysis in terms of energy and power exchange for our closed-loop controlled system. The following derivation focuses on the simplified Cartesian closed-loop system presented in \eqref {eq:closed_pos_final} where we assume $[\state,\dot\state,\ddot\state]=[\xpos,\dot\xpos,\ddot\xpos]$. This energy analysis also applies to the rotational terms, however, for simplicity of explanation we focus only on Cartesian.

To analyze the stability and passivity of our system defined in Eq. \eqref {eq:closed_pos_final}, we begin by proposing the following storage function including kinetic and potential energy,  
\begin{equation}
\label{eq:storage}
\work(\state, \dot\state) = \frac{1}{2}
  \dot\state^T \mass \dot\state + \frac{1}{2}(\state - \est\state^*)^T\mathbf{K}_p^h(c_p(t))(\state - \est\state^*)
\end{equation}
the first term in our storage function corresponds to the kinetic energy of our couple object-robot system and the second term is a potential energy generated by the estimated Cartesian DS parameters (Eq. \eqref{eq: Apest}) converging to $ \est\state^*$ and shaped by $\mathbf{K}_p^h(c_p(t))$ from Eq.~\eqref{eq:apparent_stiffness}. Eq.~\eqref{eq:storage} accounts for the kinetic energy produced by translational motion and a confidence-based varying potential function corresponding to the error that is explicitly being minimized in the tracked Cartesian DS; i.e., $\lim_{t\rightarrow\infty}||\state - \est\state^*||_2= 0$. 

The rate of change of Eq.~\eqref{eq:storage} is, 
\begin{equation}
\label{eq:storage_deriv}
\begin{aligned}
\dot\work(\state, \dot\state) & = \dot\state^T\mass\ddot\state + \frac{1}{2}
  \dot\state^T \dot\mass \dot\state + \dot\state^T\mathbf{K}_p^h(c_p(t))(\state - \est\state^*) + ...\\
  & ~~~ + \frac{1}{2}(\state - \est\state^*)^T\mathbf{\dot{K}}_p^h(c_p(t))(\state - \est\state^*)
\end{aligned}
\end{equation}
Substituting $\mass\ddot\state$ from Eq.~\eqref{eq:closed_pos_final} and using the skew-symmetry property of $\dot{\mass}-2\cori$ then $\dot\work(\state, \dot\state)$ reduced to, 
\begin{equation}
\label{eq:storage_deriv_sub}
\begin{aligned}
\dot\work(\state, \dot\state) & = \dot\state^T\uh' - \dot\state^T\impD_{p}(c_{p}(t))\dot\state + \dot\state^T\impD_{p}(c_{p}(t))\dsf_\pos(\state) + ...\\
& ~~~~ + \dot\state^T\mathbf{K}_p^h(c_p(t))(\state - \est\state^*) + ...\\
  & ~~~~ + \frac{1}{2}(\state - \est\state^*)^T\mathbf{\dot{K}}_p^h(c_p(t))(\state - \est\state^*)
\end{aligned}
\end{equation}
With $\mathbf{D}_p(c_p(t)) = c_{p}(t) \mathbf{\Lambda}_p$,  $\mathbf{K}_p^h(c_p(t)) = -c_{p}(t) \mathbf{\Lambda}_p \est\dsAp$ and $\mathbf{\dot{K}}_p^h(c_p(t)) = -\dot{c}_{p}(t) \mathbf{\Lambda}_p \est\dsAp$ and leveraging Eq.~\eqref{eq: Apest}, Eq.~\eqref{eq:storage_deriv_sub} can then be re-written as, 
\begin{equation}
\label{eq:storage_deriv_final}
\begin{aligned}
\dot\work(\state, \dot\state) & = \dot\state^T\uh' - c_{p}(t) \dot\state^T \mathbf{\Lambda}_p\dot\state + ... \\
& ~~~~ \hcancel[green]{+ \dot\state^T(c_{p}(t)\mathbf{\Lambda}_p\est\dsAp)(\state - \est\state^*)} + \dots \\
& ~~~~ \hcancel[green]{-\dot\state^T(c_{p}(t) \mathbf{\Lambda}_p \est\dsAp)(\state - \est\state^*)} + \dots \\
  & ~~~~ - \frac{\dot{c}_p(t)}{2}(\state - \est\state^*)^T\mathbf{\Lambda}_p \est\dsAp(\state - \est\state^*)\\
  &= \dot\state^T\uh' - c_{p}(t) \dot\state^T \mathbf{\Lambda}_p\dot\state \\
  & ~~~ - \frac{\dot{c}_p(t)}{2}(\state - \est\state^*)^T\mathbf{\Lambda}_p \est\dsAp(\state - \est\state^*)
\end{aligned}
\end{equation}

\noindent \textbf{Stability when $\uh' = \zerovec$:} In the absence of human guidance, when $\uh' = \zerovec$, the power in the closed-loop system is, 
\begin{equation}
\label{eq:power_full_nohuman}
\dot{\work} = - \underbrace{c_{p}(t)}_{>0} \underbrace{\dot\state^T\mathbf{\Lambda}_p\dot\state}_{> 0} - \frac{\dot{c}_p(t)}{2}\underbrace{(\state - \est\state^*)^T\mathbf{\Lambda}_p \est\dsAp(\state - \est\state^*)}_{<0}
\end{equation}
As shown, stability of the closed-loop system depends on $c_{p}(t)$ and $\dot{c}_{p}(t)$, following we analyze the different cases. 
\begin{itemize}
\item When $c_{p}(t)= 0$ then $\dot\work=\zerovec$ there is no power exchange and the robot is purely compensating for gravity.
\item When $c_{p}(t)= 1$ then the robot is following the desired velocity and only the dissipative energy term appears, 
\begin{equation}
\dot{\work} = - c_{p}(t) \dot\state^T\mathbf{\Lambda}_p\dot\state < 0
\end{equation}
which is negative definite. Which means that when the robot is very confident about the estimates $c_{p}(t) = 1$ then the closed-loop system is stable. 
\item When $0 < c_{p}(t) \leq 1$ and $\dot{c}_{p}(t)<0$, then clearly $\dot\work< 0$. Which means that when the confidence is decreasing, $c_{p}(t)\downarrow$ due to $e_p(t) \uparrow$, then the closed-loop system is stable as the potential energy is being dissipated. 

\item When $0 < c_{p}(t) < 1$ and $\dot{c}_{p}(t)>0$, then the system is stable, $\dot\work< 0$, iff the following inequality holds, 
\begin{equation}
\label{eq:confidence_bound_stability}
    - \underbrace{c_{p}(t)}_{>0} \underbrace{\dot\state^T\mathbf{\Lambda}_p\dot\state}_{> 0} \leq \frac{\dot{c}_p(t)}{2}\underbrace{(\state - \est\state^*)^T\mathbf{\Lambda}_p \est\dsAp(\state - \est\state^*)}_{<0}
\end{equation}
As shown in Eq. \eqref{eq:confidence_dt} and Eq. \eqref{eq:dp_bound_passivity} from the main text such inequality is satisfied when, 
\begin{equation}
\label{eq:app_bound}
\begin{aligned}
 -2c_{p}(t)\frac{\dot\state^T\mathbf{\Lambda}_p\dot\state}{(\state - \est\state^*)^T\mathbf{\Lambda}_p \est\dsAp(\state - \est\state^*)} & \leq \underbrace{\decay_{p} - e_{p}(t)}_{\dot{c}_{p}(t)}.
\end{aligned}
\end{equation}
When $\uh' = \zerovec$ the only reason the robot will deviate from the desired velocity is when a joint-space constraint (such as self-collision or joint limits) is being violated. Hence, assuming perfect tracking and joint-space feasibility, the robot is stable. However, when there is an infeasibility such as joint limits or collision with obstacle not considered in the control, then we do inject energy but than energy injected is bounded. 

\textbf{Bound on energy injected}: When $\uh' = \zerovec$, since $c_p$ is bounded with a maximum value of 1, we can compute the maximum amount of energy injected each time the confidence increases from 0 to 1. Assuming $\dot{c}_p > 0$ is constant, we have
\begin{align}
-\int_0^\frac{1}{\dot c_p} \frac{\dot c_p}{2}(\state - \est\state^*)^T\mathbf{\Lambda}_p \est\dsAp(\state - \est\state^*) ds =\\
-\frac{1}{2}(\state - \est\state^*)^T\mathbf{\Lambda}_p \est\dsAp(\state - \est\state^*)
\end{align}
which can be shown to be bounded.
\end{itemize}

\begin{figure}[!tbp]
    \centering
    \includegraphics[trim={1.2cm 0 1.2cm 1.2cm},clip, width=0.495\textwidth]{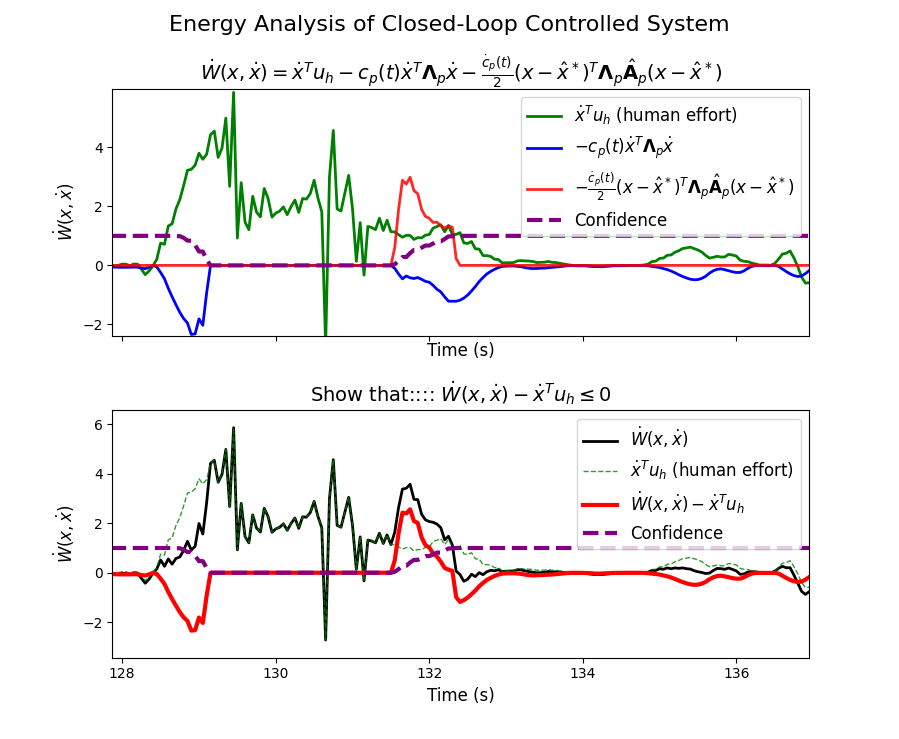}
    \caption{\revise{An instance where the third term of Eq. \eqref{eq:storage_deriv_final} is positive for a short period of time when confidence is increasing, and Eq. \eqref{eq:passivity_condition} does not hold. \textbf{top:} energy terms of Eq. \eqref{eq:storage_deriv_final}, and \textbf{bottom:} analysis of Eq. \eqref{eq:passivity_condition} for the proposed method.}}
    \label{fig: completion_force}
    \vspace{-10pt}
\end{figure}

\noindent \textbf{Passivity when $\uh' \neq \zerovec$:} To examine the stability properties of the closed-loop dynamics of the proposed system while interacting with the human, i.e., $\uh' \neq \zerovec$, we use the concept of passivity which focuses on analyzing the energy dissipation of the system as in \cite{passiveDS,khoramshahi2019dynamical,khoramshahi2020dynamical}. In this case, we seek to analyze the passivity of the mapping of the human force to robot velocity $\uh' \rightarrow \dot\state$. Hence, to prove the passivity of the system wrt. the input-output pair $(\uh',\dot\state)$ the following condition must hold, 
\begin{equation}
\label{eq:passivity_condition}
 \mathbf{\dot{W}}(\state, \dot\state) \leq    \dot\state\T \uh'
\end{equation}
Plugging Eq.~\eqref{eq:storage_deriv_final} into this condition, 
\begin{equation}
\dot\state^T\uh' -c_{p}(t) \dot\state^T \mathbf{\Lambda}_p\dot\state  - \frac{\dot{c}_p(t)}{2}(\state - \est\state^*)^T\mathbf{\Lambda}_p \est\dsAp(\state - \est\state^*) \leq    \dot\state\T \uh'
\end{equation}
similarly to the case when $\uh' = \zerovec$, the closed-loop system is passive wrt. the input/output pair $(\uh',\dot\state)$ when $c_p(t) = 0$, $c_p(t) = 1$ and for $0 < c_p(t) < 1$ when the condition introduced in Eq.~\eqref{eq:confidence_bound_stability} holds; i.e., if the second and third term on the left hand side corresponding to the power introduced by the variable impedance control law is negative.  Otherwise, an increase in confidence corresponds to injecting potential energy in the system, which is unavoidable as noted by \cite{kronander2016stability}, unless you introduce a controller that actively compensates for this. This effect can be seen in Fig. \ref{fig: completion_force}, as shown for $d_p=0.41$ the robot looses passivity for 0.75s while the confidence is increasing and for the rest of the interaction the robot remains passive. As highlighted in the main text and in the previous section one can tune $d_p$ following Eq. \eqref{eq:app_bound} to achieve a user preferred stiffening behavior during physical interaction.

\end{document}